\documentclass[10pt,letterpaper]{article}

\usepackage{epsfig}
\usepackage{graphicx}
\usepackage{url}
\usepackage{subfigure}
\usepackage{mathtools}
\usepackage{algpseudocode}
\usepackage{algorithm}
\usepackage{amsmath}
\usepackage{relsize}
\usepackage{mathrsfs}
\usepackage{bm}
\usepackage{amsfonts}
\usepackage{color}
\usepackage{xcolor}
\usepackage{verbatim}
\usepackage{subfig}

\newcommand{\ba}{\mathbf{a}}

\newcommand{\bk}{\mathbf{k}}
\newcommand{\bx}{\mathbf{x}}
\newcommand{\bX}{\mathbf{X}}

\newcommand{\by}{\mathbf{y}}
\newcommand{\bY}{\mathbf{Y}}
\newcommand{\bI}{\mathbf{I}}
\newcommand{\bK}{\mathbf{K}}

\newcommand{\btheta}{\boldsymbol{\theta}}
\newcommand{\bpsi}{\boldsymbol{\psi}}

\newcommand{\bxi}{\boldsymbol{\xi}}

\newcommand{\Q}{Q}

\newcommand{\E}{\mathbb{E}}
\newcommand{\V}{\mathbb{V}}

\newcommand{\GP}{\operatorname{GP}}

\newcommand{\Bsum}{\mathlarger{\sum}}

\newcommand{\calN}{\mathcal{N}}

\newcommand{\sref}[1]{Sec.~\ref{sec:#1}}
\newcommand{\qref}[1]{Eq.~(\ref{eqn:#1})}

\newcommand{\fref}[1]{Fig.~\ref{fig:#1}}

\definecolor{Gray}{gray}{.25}

\begin{document}
\vspace*{0.1in}

\begin{center}
{\Large
\textbf{Learning Arbitrary Quantities of Interest from Expensive Black-Box Functions through Bayesian Sequential Optimal Design}
}
\newline

Piyush Pandita\textsuperscript{1, *},
Nimish Awalgaonkar\textsuperscript{2},
Ilias Bilionis\textsuperscript{2},
Jitesh Panchal\textsuperscript{2},
\\
\bigskip
{1} Probabilistics and Optimization Group, GE Research, Niskayuna, NY 12309
\\
{2} School of Mechanical Engineering, Purdue University, West Lafayette, Indiana 47907
\\
\bigskip
* piyush.pandita@ge.com

\end{center}

\begin{abstract}
Estimating arbitrary quantities of interest (QoIs) that are non-linear operators of complex, expensive-to-evaluate, black-box functions is a challenging problem due to missing domain knowledge and finite information acquisition budgets.
Bayesian optimal design of experiments (BODE) is a family of methods that identify an optimal design of experiments (DOE) under different contexts, such as learning a response surface, estimating a statistical expectation, solving an optimization problem, etc., using only in a limited number of function evaluations.
Under BODE methods, sequential design of experiments (SDOE) accomplishes this task by selecting an optimal sequence of experiments.
SDOE methods use data-driven probabilistic surrogate models for emulating the expensive black-box function.
Probabilistic predictions from the surrogate model are used to define an information acquisition function (IAF) which quantifies the marginal value contributed or the expected information gained by a hypothetical experiment. 
The next experiment is selected by maximizing the IAF.
A generally applicable IAF is the expected information gain (EIG) about a QoI as captured by the expectation of the Kullback-Leibler divergence between the predictive distribution of the QoI after doing a hypothetical experiment and the current predictive distribution about the same QoI.
Despite its intuitive appeal, applications of EIG have been limited due to the difficulty associated with estimating it.

To overcome this barrier, in this work, we develop a practical numerical estimator of the EIG for arbitrary QoIs.
We model the underlying information source as a fully-Bayesian, non-stationary Gaussian process (FBNSGP), derive an approximation of the information gain of a hypothetical experiment about an arbitrary QoI conditional on the hyper-parameters, and estimate the EIG about the same QoI using sampling averages to integrate over the posterior of the hyper-parameters and the potential experimental outcomes.
We demonstrate the performance of our method in four numerical examples and a practical engineering problem of steel wire manufacturing.
The method is compared to two classic SDOE methods: random sampling and uncertainty sampling.
The results show a mixed outcome where the proposed methods converges sooner with increasing dimensionality for many examples.
We publish an implementation of our work named AdaptiveBODE\footnote{to be released upon acceptance of the paper} in the Python programming language.\\

\end{abstract}

\section{Introduction}
\label{sec:intro-bode-generic}
Engineers and scientists use simulated (computer codes)~\cite{sacks1989} or physical (laboratory) experiments~\cite{flournoy1993} to learn about real-world physical phenomena. 
Even though such information sources are becoming more precise, e.g., through the development of more sophisticated mathematical models or advanced laboratory equipment, respectively, the time and monetary cost of information queries remains significant. 
On one hand, high fidelity simulations typically require super-computing facilities which have a high upfront purchasing cost as well as a large operational energy consumption footprint.
On the other hand, physical experiments may require the destruction of an expensive engineered artifact; How many times can one test a full scale aircraft in a wind tunnel?
As a result, due to the ``expensive'' nature of these information sources, the number of experiments is severely restricted~\cite{ginsbourger2010kriging}.

Not all experimental data are equally useful. 
That is, some data can provide valuable information about the quantities of interest (QoIs) while others may be irrelevant to the goals of an investigation~\cite{huan2016sequential}. 
The challenge is to design experiments that optimally allocate one's information acquisition budget for learning a specific QoI. 
For example, the researcher might be interested in augmenting their state-of-knowledge about the response surface of a QoI, or just its maximum value~\cite{sacks1989,hennig2012entropy,contal2014gaussian}, its statistical expectation or variance, or other percentiles~\cite{oakley2004estimating}.
Experimental design addresses the questions of which information source to query, where to query it, and when to stop querying it in order to achieve one's learning goals~\cite{huan2016sequential}.

The opportunities arising from the development of systematic design of experiments strategies has generated a lot of interest within the science and engineering community~\cite{brochu2010,snoek2012practical}. 
In the context of expensive black-box simulators, traditional design of experiments (DOE) methods usually face three major hurdles: 
a) they require hundreds of thousands of expensive simulations, 
b) they may require gradients of the simulator, and
c) they do not take into account/exploit knowledge of the underlying physical process.

Bayesian optimal design of experiments (BODE) overcomes these obstacles.
The key characteristics of BODE approaches are that:
a) they maintain a probabilistic representation of the underlying response surface, e.g., using Gaussian process regression (GPs);
b) in a low data regime, by strictly adhering to a Bayesian paradigm, they handle uncertainty in a principled manner;
c) they incorporate prior knowledge about the response surface from different sources, e.g., that the lengthscale of the response dependent on input variables~\cite{heinonen2016non}, or that the response is uni-modal in nature~\cite{andersen2017bayesian}, etc.

Among BODE methods, Bayesian sequential design of experiments (BSDOE) strategies \cite{locatelli1997bayesian,jones1998,gaul2014modified,huang2006global,lizotte2008,frazier2008knowledge,feliot2018user} which use information acquisition functions (IAF) like the probability of improvement (PI), the expected improvement (EI), the knowledge gradient (KG),  etc., allow for new experiments to be chosen and conducted in a sequential manner, thus using newly acquired data to guide the design of future experiments~\cite{huan2016sequential}.
A conventional BSDOE is essentially a problem of sequential decision-making under uncertainty~\cite{huan2016sequential}, which  involves the following four main steps:
a) construct a probabilistic surrogate model for characterizing our state of knowledge about the expensive information source (prior predictive distribution);
b) condition this probabilistic surrogate model based on the observed data (posterior predictive distribution);
c) use the newly inferred posterior predictive distribution to compute the (user-defined) IAF values for hypothetical experiments;
d) the hypothetical experiment which maximizes the IAF function is chosen as the new experiment to run next.
One keeps repeating steps b to d until they exhaust the available budget or a predefined stopping criterion is met.

The choice of the IAF plays a crucial role in the efficacy of the BSDOE algorithms since it encodes the objectives of the investigator.
For example, the expected improvement (EI) \cite{viana2013efficient, bull2011, vazquez2010convergence} is appropriate when the investigator is interested in optimizing the information source with respect to a controllable parameter.
In general, a new IAF needs to be designed for every new quantity of interest (QoI).
According to information theory, it is possible to define the expected information gain about an arbitrary QoI as the expectation of the Kullback-Leibler
divergence~\cite{kullback1997information} (KLD) between the predictive distribution of the QoI after doing a hypothetical experiment and the current predictive distribution about the same QoI.
For example, this idea is used in uncertainty sampling (US)~\cite{mackay1992information} which is a BODE heuristic derived from maximizing the expected information gain about the parameters of the probabilistic surrogate, i.e., the implicit goal of US is to learn the entire response surface.
Other examples include the sensor placement problem~\cite{nath2017sensor,huan2014gradient,lin2019approximate},
surrogate modeling~\cite{yan2018gaussian,choi2004polynomial,hadigol2017least},
learning missing parameters~\cite{terejanu2013bayesian},
optimizing an expensive physical response~\cite{hennig2012entropy},
calibrating a physical model~\cite{guestrin2005near,huan2013simulation},
reliability design~\cite{picheny2010adaptive},
efficient design space exploration~\cite{liu2016adaptive},
probabilistic sensitivity analysis~\cite{liu2006relative},
portfolio optimization~\cite{gonzalvez2019financial},
hyperparameter tuning~\cite{wu2019practical},
human experiment design~\cite{foster2019variational},
radiation detector placement~\cite{michaud2019simulation}, and
estimation of statistical expectation~\cite{pandita2019bayesian}.

Even though EIG is, in principle, generally applicable, each new QoI requires considerable theoretical developments to derive a suitable numerical approximation.
To overcome this barrier, the \emph{objective of this paper} is to develop a practical numerical estimator of the EIG for arbitrary QoIs.
Throughout this paper, this estimate of the EIG is referred to as the expected KLD (EKLD).

The performance of BSDOE algorithms also depends in a crucial way on the choice of the probabilistic surrogate model used to represent the underlying information source.
A common assumption made by such surrogates, like Gaussian process regression (GPR)~\cite{o2006bayesian}, about that the information source has constant smoothness and constant signal variance across the input domain.
This assumption can have ill-effects on the modeling and the subsequent sequential design process when the underlying function has discontinuities or sharp changes in smoothness as shown in~\cite{heinonen2016non,mohammadi2019emulating}. 
To tackle this problem, we introduce a fully-Bayesian non-stationary Gaussian process (FBNSGP) \cite{gibbs1998bayesian,paciorek2003nonstationary,heinonen2016non,mohammadi2019emulating} surrogate model to quantify our beliefs about the objective function. FBNSGP, where the two key parameters - signal variance and lengthscale - can be simultaneously input dependent, helps us to model realistic input-dependent dynamics~\cite{heinonen2016non}.
Furthermore, this surrogate model allows the scientist's expertise to be incorporated, as prior knowledge, in the surrogate model at the highest hierarchical level of the parameterization.
Previous work done in developing non-stationary emulators for such problems include the treed-GP model of Grammacy et al.~\cite{gramacy2008bayesian}, the GP-experts based model of Rasmussen et al.~\cite{rasmussen2002infinite}, point estimates of local smoothness based non-stationary GP modeling by~\cite{paciorek2004nonstationary} and~\cite{plagemann2008nonstationary}.

To summarize, the contributions of this work are as follows:
a) Firstly, we examine the performance of the EKLD in the context of BSDOE for arbitrary QoIs that are non-linear operators of an expensive information source.
Specific examples of QoIs include, the statistical expectation and variance, percentiles, the maximum, and the minimum.
We perform experiments on numerical examples with varying characteristics such as the number of input dimensions, initial samples, smoothness, etc.
We also illustrate the performance of the developed framework on an engineering steel wire manufacturing problem;
b) Secondly, we compare and contrast the convergence of the EKLD to state-of-the-art BSDOE methods.
c) Thirdly, we publish an implementation of our work named AdaptiveBSDOE\footnote{https://github.com/piyushpandita92/} as an open-source
package in the Python language.

The remainder of the paper is organized as follows.
This paper starts with an introduction of problem at hand, BSDOE methodology used for addressing this problem etc. in \sref{metho-bode-generic}.
In \sref{results-bode-generic}, the performance of the newly developed KLD-FBNSGP based BSDOE framework is evaluated (by showing how this framework would work in case of synthetic examples).
In \sref{comparison-bode-generic}, comparison of our methodology with state-of-the-art BSDOE methods namely, US and EI, is presented in different contexts.
\sref{conclusions_bode_generic} summarizes our findings and concludes the paper.

\section{Methodology}
\label{sec:metho-bode-generic}

Let $\mathcal{X}$ be a compact, connected subset of $\mathbb{R}^d$ and $f:\mathcal{X}\rightarrow \mathbb{R}$ a bounded, square integrable function.
We refer to $\mathcal{X}$ as the \emph{experimental design space} and to $f$ as the \emph{information source}.
A point $\bx$ in $\mathcal{X}$ corresponds to an \emph{experimental condition} (or \emph{experimental input} or simply \emph{input}).
An experiment $\bx$ results in a \emph{measurement} $y$ of the hidden \emph{information source} $f(\bx)$.
The likelihood of $y$ conditioned on $\bx$ and $f(\bx)$ depends on the, potentially unknown, details of the measurement process, see Sec.~\ref{sec:likelihood}.

A QoI, denoted $Q[f]$, is any non-linear functional of $f$.
Examples of QoIs are the maximum and the minimum of $f$ defined by:
\begin{equation}
    \max[f] := \max_{\bx\in\mathcal{X}}f(\bx),
\end{equation}
and
\begin{equation}
    \min[f] := \min_{\bx\in\mathcal{X}}f(\bx),
\end{equation}
respectively.
To provide some examples of statistical QoIs, let $(\Omega,\mathcal{F},\mathbb{P})$ be a probability space and $\mathbf{X}$ a random vector (r.v.) with support $\mathcal{X}$.
The statistical expectation and variance of $f$, defined by
\begin{equation}
    \label{eqn:qoi_ex}
    \mathbb{E}[f] := \mathbb{E}\left[f(\mathbf{X})\right],
\end{equation}
and 
\begin{equation}
    \mathbb{V}[f] := \mathbb{V}\left[f(\mathbf{\mathbf{X}})\right] := \mathbb{E}\left[\left(f(\mathbf{X}) - \mathbb{E}[f]\right)^2\right],
\end{equation}
respectively, are QoIs.
Finally, another QoI of interest is the $\alpha$-percentile of $f$:

\begin{equation}
    \mathbb{Q}_\alpha[f] := \inf\left  \{y: y\in\mathbb{R}, \mathbb{P}[f(\mathbf{X}) \le y] \geq \alpha \right\}.
\end{equation}

Our problem is to learn a given QoI $Q[f]$ by sequentially selecting a finite number $N$ of experiments.
At the $n$-th step of the BSDOE algorithm, we use the selected experimental inputs $\bX_n = (\bx_1,\dots,\bx_n)$ and the corresponding observations $\by_n = (y_1,\dots,y_n)$ to update our beliefs about $f$ in a Bayesian manner~\cite{jaynes2003probability}, quantifying the epistemic uncertainty induced by limited number of data (Sec.~\ref{sec:qoi_belief}).
Then, we select the new experiment to run by maximizing the expected information gain for the quantity of interest $\Q[f]$ (Sec.~\ref{sec:ekld}).

\subsection{Quantifying our state of knowledge about the underlying information source}
\label{sec:qoi_belief}
We learn $f$ from data using GP regression~\cite{rasmussen2006,mackay1998introduction}, a commonly used class of non-parametric probabilistic surrogate modeling techniques.
GP surrogate models combine prior knowledge with data to define a posterior distribution over the function space of interest.
The Bayesian nature of the method enables one    to quantify the epistemic uncertainty induced by limited data which is essential for defining the expected information gain of any QoI.

A GP requires the selection of a prior mean function $m:\mathcal{X}\rightarrow \mathbb{R}$ and a positive-definite covariance function $k:\mathcal{X}\times\mathcal{X}\rightarrow\mathbb{R}$.
Without loss of generality we pick the mean to be identically zero.
The covariance function $k(\bx,\bx')$ defines the a priori correlation between the responses $f(\bx)$ and $f(\bx')$.
Standard GP approaches assume a stationary covariance, i.e., $k(\bx,\bx') = k(\bx - \bx')$.

However, this strong assumption is not justified in many real-world engineering problems where either the response surface may exhibit an input-dependent variance or lengthscale, or both~\cite{heinonen2016non}.

Several works have suggested solutions to observed non-stationarity.
Recent examples of such non-stationary GPs (NSGP) include the work of~\cite{binois2018practical} on heteroscedastic (input-dependent) noise~\cite{binois2018practical}, and the local lengthscale modeling of~\cite{plagemann2008nonstationary,kersting2007most,paciorek2004nonstationary}.
These NSGP models follow a hierarchical Bayes approach.
First, they model the underlying information source as a GP, albeit with a modified covariance function with explicitly input-dependent hyper-parameters.
Second, they assign GP priors to the covariance hyper-parameters.
Unfortunately, the posterior of the NSGP becomes analytically intractable~\cite{tolvanen2014expectation} and it can only be characterized via sampling.
To this end, in this work we perform fully-Bayesian inference using Hamiltonian Monte Carlo (HMC)~\cite{duane1987hybrid,neal2011mcmc,matthews2017gpflow} to generate samples from the posterior distribution of the hyperparameters of the NSGP model. 
This NSGP model is implemented using the GPflow 0.4.0 package~\cite{GPflow2017},
a GP library that allows one to implement GP models using tensorflow. 
The following sections have the necessary details pertaining to our NSGP model.

\subsubsection{Modeling prior beliefs about the information source}
\label{sec:prior_gen}
Prior to seeing any data, we model our beliefs about $f$ as a zero mean GP with covariance function $k$, i.e.,
\begin{equation}
    \label{eqn:prior}
    f | \mathbf{s}, \mathbf{l} \sim \GP(0, k(\cdot,\cdot;\mathbf{s},\mathbf{l}),
\end{equation}
where $k$ is a non-stationary covariance function, first suggested by Gibbs~\cite{gibbs1998bayesian} and also used in~\cite{plagemann2008nonstationary,paciorek2004nonstationary,paciorek2006spatial},
\begin{equation}
    \label{eqn:prior_cov_func}
    k\left(\bx, \bx';\mathbf{s}, \mathbf{l}\right) =  \prod_{i=1}^{d}s_{i}(x_{i})s_{i}(x'_{i})\sqrt{\frac{l_{i}(x_{i})l_{i}(x'_{i})}{l^{2}_{i}(x_{i}) + l^{2}_{i}(x'_{i})}}\exp{\left\{-\frac{(x_{i}-x'_{i})^{2}}{l^{2}_{i}(x_{i}) + l^{2}_{i}(x'_{i})}\right\}},
\end{equation}
for any $\bx$ and $\bx'$ in $\mathcal{X}$, with $\mathbf{s} = (s_1,\dots,s_d)$, $\mathbf{l} = (l_1,\dots,l_d)$ with $s_{i}:\mathbb{R}\rightarrow\mathbb{R}$ and $l_{i}:\mathbb{R}\rightarrow\mathbb{R}$ being positive, input-dependent signal-variance and lengthscale functions corresponding to input dimension $i$, respectively.
The dimensionality of the input space is denoted by $d$.
Note that, the covariance function defined in \qref{prior_cov_func} is positive definite and reduces into a standard stationary squared exponential covariance if both the signal-variance and lengthscale are assumed to be constant.
We model the signal-variances $s_{i}$ and the lengthscales $l_{i}$  using latent GPs.
To ensure their positivity, we place separate GP priors on the logarithms of $s_{i}$ and $l_{i}$, i.e.,
\begin{equation}
    \label{eqn:prior_ss}
    \log (s_{i}) | \bpsi_{s,i} \sim \GP(m_{s, i}, k_{s, i})
\end{equation}
and, 
\begin{equation}
    \label{eqn:prior_ell}
    \log (l_{i}) | \bpsi_{l,i}\sim \GP(m_{l, i}, k_{l, i}),
\end{equation}
where the means $m_{\lambda,i}$ are just constants, and the covariance functions $k_{\lambda,i}(x_i, x_i')$ are squared exponentials:
\begin{equation}
    \label{eqn:prior_cov_func_lat}
    k_{\lambda,i}(x_i, x_i'; \bpsi_{\lambda,i}) = v^{2}_{\lambda,i}\exp\left\{-\frac{(x_i-x_i')^2}{2\ell_{\lambda,i}^2}\right\},
\end{equation}
for $\lambda = s$ or $l$.
That is, each latent GP has three hyperparameters $\bpsi_{\lambda,i} = \left(m_{\lambda,i}, v_{\lambda,i}, \ell_{\lambda,i}\right)$.
Denote all of the hyperparameters collectively by $\bpsi = \left( (\bpsi_{s, 1}, \bpsi_{l, i}), \cdots, (\bpsi_{s, d}, \bpsi_{l, d})\right)$ and assume that all these hyperparameters are a priori independent:
\begin{equation}
    p(\bpsi) = \prod_{\lambda = s,l}\prod_{i=1}^dp(m_{\lambda,i})p(v_{\lambda,i})p(\ell_{\lambda,i}).
\end{equation}
We place a Gamma prior with unit parameters, $\mathcal{G} (1, 1)$, on the lengthscales $\ell_{\lambda,i}$ and the signal-variances $v_{\lambda,i}$.
The priors for the means $m_{\lambda,i}$ are selected differently for one-dimensional and multi-dimensional problems and will be highlighted in the numerical examples.

All computations are necessarily performed on a discrete set of inputs.
Corresponding to the observed inputs, define $\mathbf{f}_n = \left(f(\bx_1),\dots,f(\bx_n)\right)$ to be the vector of function values and $\mathbf{l}_{i,n} = \left(l_i(\bx_1),\dots,l_i(\bx_n)\right)$ and $\mathbf{s}_{i,n} = \left(s_i(\bx_1),\dots,s_i(\bx_n)\right)$ to be the vectors of lengthscales and signal variances of each dimension, respectively.
For notational convenience, we write collectively $\mathbf{L}_n = \left(\mathbf{l}_1,\dots,\mathbf{l}_d\right)$ and $\mathbf{S}_n = \left(\mathbf{s}_1,\dots,\mathbf{s}_n\right)$.
On this discretization, the previously defined GPs induce the following prior:
\begin{equation}
    p(\mathbf{f}_n, \mathbf{L}_n, \mathbf{S}_n|\bX_n, \bpsi) = p\left(\mathbf{f}_n|\bX_n, \mathbf{L}_n,\mathbf{S}_n\right)\prod_{i=1}^d p\left(\mathbf{l}_{i,n}|\bX_n,\bpsi_{l,i}\right)p\left(\mathbf{s}_{i,n} |\bX_n,\bpsi_{s,i}\right),
\end{equation}
where all the terms are multivariate Gaussians with mean vectors and covariance matrices induced by the previously defined GPs.

\subsubsection{Modeling the measurement process}
\label{sec:likelihood}

The likelihood function is a model of the measurement process, i.e., it establishes the connection between the information source output $f(\bx)$ and the observed data $y$.
In this work, we assume an additive noise model with input-independent noise (extending to heteroscedastic noise can be achieved using latent GPs as in the previous section).
The data-likelihood is:
\begin{equation}
    \label{eqn:likelihood}
    p(\by_{n} | \mathbf{f}_n, \sigma) = \mathcal{N}(\by_{n}|\mathbf{f}_n, \sigma^{2}\mathbf{I}_{n}),
\end{equation}
where $\mathbf{I}_n$ is the $n\times n$ identity matrix. We fix the noise-variance $\sigma^{2}$ equal to 1e-6 for all our problems in this paper.

\subsubsection{Bayesian inference}
\label{sec:post_theta}
The posterior of all latent GPs and the corresponding hyper-parameters is given by:
\begin{equation}
    \label{eqn:post_gp_full}
    p(\mathbf{f}_n, \mathbf{L}_n, \mathbf{S}_n, \bpsi|\bX_n, \by_n,\sigma) \propto p(\by_n|\mathbf{f}_n,\sigma)p(\mathbf{f}_n, \mathbf{L}_n, \mathbf{S}_n|\bX_n,\bpsi)p(\bpsi),
\end{equation}
where all terms have been previously defined.
As discussed earlier, we use HMC to obtain $M$ posterior samples $\mathbf{f}_{n,m}, \mathbf{L}_{n,m}, \mathbf{S}_{n,m}, \bpsi_m, m=1,\dots,M$.

\subsubsection{Making predictions}
\label{sec:post_gp_gen}
In order to be able to make predictions at an unobserved input, one needs to characterize the aposteriori state-of-knowledge about $f$ conditioned on the hyperparameters and the data.
Instead of sampling from the joint posterior \qref{post_gp_full} we sample from the posterior of $f$ conditioned on a sample from the posterior of lengthscales and the signal-variances from the respective latent GPs.
The conditional posterior of $f$ is characterized by the following GP:
\begin{equation}
    \label{eqn:post_gp_gen}
    f|\mathbf{X}_n, \mathbf{y}_n,\mathbf{S}_{n,m}, \mathbf{L}_{n,m}, \bpsi_m \sim \GP(f|w_{n,m}, k_{n,m}),
\end{equation}
with \emph{conditional posterior mean} function:
\begin{equation}
    \label{eqn:posterior_mean}
    w_{n,m}(\bx) = \left(\bk_{n,m}(\bx)\right)^{T}\left(\bK_{n,m} + \sigma^2\bI_n\right)^{-1}\mathbf{y}_{n},
\end{equation}
where $\bK_{n,m}$ is the covariance matrix, $K_{n,m,ij} = k(\bx_i,\bx_j)$, the term $\bk_{n,m}(\bx) = \left(k(\bx,\bx_1), \dots, k(\bx,\bx_n)\right)^T$ is the cross-covariance, and \emph{conditional posterior covariance} function is:
\begin{equation}
    \label{eqn:predictive_covariance}
    k_{n,m}(\bx,\bx') = k(\bx, \bx') - \left(\bk_{n,m}(\bx)\right)^{T}\left(\bK_{n,m} + \sigma^2\bI_n\right)^{-1} \bk_{n,m}(\bx').
\end{equation}
The values of lengthscale and signal-strength at $\tilde{\bx}$ for each sample $m$, namely $l_{i,m}(\tilde{\bx})$ and $s_{i,m}(\tilde{\bx})$, are approximated as the mean of their respective latent GPs, conditioned on $\bpsi_m, \mathbf{L}_{n,m}$.
In particular, at an untried test input $\tilde{\bx}$ the point-predictive posterior probability density  of $\tilde{y} = f(\tilde{\bx})$ conditioned on the hyperparameters is:
\begin{equation}
    \label{eqn:point_predictive}
    p(\tilde{y}|\tilde{\bx}, \mathbf{X}_n, \mathbf{y}_n,\mathbf{S}_{n, m}, \mathbf{L}_{n, m}, \bpsi_m) = \calN\left(\tilde{y}\middle|w_{n,m}(\tilde{\bx}), \sigma_{n,m}^2(\tilde{\bx})\right)
\end{equation}
where $\sigma_{n,m}^2(\tilde{\bx}) = k_{n,m}(\tilde{\bx},\tilde{\bx})$.
\subsection{Karhunen-Lo\`eve expansion of the conditional NSGP}
As mentioned earlier, the ultimate goal is to characterize our state of knowledge about QoIs that are functionals of $f$, i.e., $Q=Q[f]$.
To this end, we need to be able to sample analytical functions $f_{n,m}(\tilde{\bx})$ from our conditional state of knowledge.
To achieve this, we employ the Karhunen-Lo\`eve expansion (KLE) of the conditional posterior of $f$ in \qref{post_gp_gen}.
The KLE~\cite{ghanem1991stochastic} of the posterior of $f$ at an input $\tilde{\bx}$ is given by:
\begin{equation}
    \label{eqn:kle_full}
    f_{n,m}(\tilde\bx; \bxi) = w_{n,m}(\tilde\bx) + \sum_{i=1}^{\infty}\xi_{i}\sqrt{\eta_{n,m,i}}\phi_{n,m,i}(\tilde\bx),
\end{equation}
where $w_{n,m}(\tilde{\bx})$ is the conditional posterior predictive mean of \qref{posterior_mean}, and the random variables $\bxi$ are independent identically distributed (iid) standard normal. 
The scalars $\eta_{n,m,i}$ and the functions $\phi_{n,m,i}(\tilde{\bx})$ are the eigenvalues and eigenfunctions of the conditional posterior covariance function given in Eq.~\qref{predictive_covariance}.
We identify these terms via the Nystr\"om approximation~\cite{reinhardt2012analysis,betz2014numerical} by retaining the eigenvalues and corresponding eigenvectors of the posterior covariance matrix constructed using a random (LHS) quadrature.
In our work, we sample 500 points for one-dimensional problems and 5000 points for multi-dimensional problems to construct the covariance matrix and obtain the eigenvalues and the eigenfunctions for the KLE.
We truncate \qref{kle_full} to a finite number of stochastic terms, given by $W$, as follows:
\begin{equation}
    \label{eqn:kle_trunc}
    f_{n,m}(\tilde\bx; \bxi) \approx w_{n,m}(\tilde\bx) + \sum_{i=1}^{W}\xi_{i}\sqrt{\eta_{n,m,i}}\phi_{n,m,i}(\tilde\bx).
\end{equation}
More details about the same can be found in Section 2.8.1 of Bilionis et. al.~\cite{bilionis2016}.
The number of terms in the KLE, $W$, is determined by specifying the percentage $\beta$ of the total sum of the eigenvalues to be retained as follows:
\begin{equation}
    \label{eqn:energy}
    \Bsum_{i=1}^{W}\eta_{n,m,i} = \beta \Bsum_{i=1}^{\infty}\eta_{n,m,i}.
\end{equation}
For our experiments we take the value of $\beta$ equal to 0.95, i.e., the truncated KLE explains 95\% of the total variance of the posterior GP.

\subsubsection{Conditioning the KLE of the conditional state-of-knowledge on a hypothetical query}

Now, consider a hypothetical observation denoted by $\tilde{y}$ at a test point (input) $\tilde{\bx}$.
Employing the KLE of \qref{kle_trunc} along with the measurement process model, the predictive distribution of $\tilde{y}$ conditioned on $\bxi$ (and on everything at the $m$-th HMC step):
\begin{equation}
    \label{eqn:est_hyp}
    p(\tilde{y}|\tilde{\bx}, \bxi, \bX_n, \by_n,\mathbf{S}_{n,m},\mathbf{L}_{n,m},\bpsi_{m},\sigma) = \mathcal{N}(\tilde{y}| f_{n,m}(\tilde{\bx};\bxi), \sigma^{2}).
\end{equation}
Deriving the posterior of $\bxi$ conditional on the $\mathbf{D}_{n}$, $\tilde{\bx}$ and $\tilde{y}$ by completing the squares, results in the following:
\begin{eqnarray}
    \label{eqn:post_xi}
    p(\bxi|\tilde{\bx}, \tilde{y}, \bX_n, \by_n,\mathbf{S}_{n,m},\mathbf{L}_{n,m},\bpsi_{m},\sigma) \propto p(\tilde{y}|\bxi, \tilde{\bx},\bX_n, \by_n,\mathbf{S}_{n,m},\mathbf{L}_{n,m},\bpsi_{m},\sigma)p(\bxi) \nonumber \\
    \Rightarrow p(\bxi|\tilde{\bx}, \tilde{y}, \bX_n, \by_n,\mathbf{S}_{n,m},\mathbf{L}_{n,m},\bpsi_{m},\sigma) = \mathcal{N}(\bxi|{\bm{\mu}}_{n,m}(\tilde\bx), {\bm{\Sigma}}_{n,m}(\tilde\bx))),\nonumber \\
\end{eqnarray}
where,
\begin{equation}
    \label{eqn:post_mean}
    {{\bm{\mu}}}_{n,m}(\tilde\bx) = {\bm{\Sigma}}_{n,m}(\tilde{\bx}){\ba_{n,m}^{T}(\tilde{\bx})}\left(\frac{\tilde{y} - w_{n,m}(\tilde{\bx})}{\sigma^{2}}\right),
\end{equation}
\begin{equation}
    \label{eqn:post_cov}
    {\bm{\Sigma}}_{n,m}(\tilde\bx) = \left[\bI_{W} + \ba_{n,m}^{T}(\tilde\bx)\ba_{n,m}(\tilde\bx)\frac{1}{\sigma^{2}}\right]^{-1},
\end{equation}
with,
\begin{eqnarray}
    \label{eqn:const_vec}
    \ba_{n,m} = \left(\sqrt{\eta_{n,m,1}}\phi_{n,m,1}(\tilde\bx), \dots, \sqrt{\eta_{n,m,W}}\phi_{n,m,W}(\tilde\bx)\right),
\end{eqnarray}
The Sherman-Morrison formula~\cite{sherman1950adjustment} allows us to express the posterior covariance of $\bxi$ from \qref{post_cov} as:
\begin{eqnarray}
    \label{eqn:sigma_p}
    {\bm{\Sigma}}_{n,m}(\tilde\bx) = \bI_{W} - \frac{\ba_{n,m}^{T}(\tilde\bx)\ba_{n,m}(\tilde\bx)}{\sigma^{2} + \ba_{n,m}(\tilde\bx)\ba_{n,m}(\tilde\bx)^{T}}.
\end{eqnarray}
Thus, an element of the posterior covariance matrix of $\bxi$ can be expressed as:
\begin{eqnarray}
    \label{eqn:sigma_p_final}
    {\bm{\Sigma}}_{n,m,ij}(\tilde\bx) = \delta_{ij} - \frac{\sqrt{\eta_{n,m,i}\eta_{n,m,j}}\phi_{n,m,i}(\tilde\bx)\phi_{n,m,j}(\tilde\bx)}{\sigma^{2} + \sum_{i=1}^{W}\eta_{n,m,i}\phi_{n,m,i}^2(\tilde\bx)},
\end{eqnarray}
where $\delta_{ij}$ is the Kronecker delta.

\subsection{Sequential design of experiments using the expected information gain}
\label{sec:ekld}
Consider the PDF of the QoI $\Q[f]$ conditional on the observed data and the $m$-th HMC sample from the posterior of all latent parameters:
\begin{equation}
    \label{eqn:current_Q}
    p_{n,m}(q) := \mathbb{E}\left[\delta\left( \Q[f] - q\right)\middle|\bX_n, \by_n, \mathbf{S}_{n,m}, \mathbf{L}_{n,m}, \bpsi_m\right],
\end{equation}
where $\delta$ is Dirac's delta function and the expectation is over the function space measure defined by the posterior GP, see Eq.~(\ref{eqn:post_gp_gen}).
The uncertainty in $p_{n,m}(q)$ represents our epistemic uncertainty induced by the limited number of observations.
Now consider the hypothetical output $\tilde{y}$ at a hypothetical experimental input $\tilde{\bx}$.
Our (hypothetical) state of knowledge about $Q[f]$ becomes: 
\begin{equation}
    \label{eqn:hypothetical_Q}
    p_{n,m}(q|\tilde{\bx}, \tilde{y}) = \mathbb{E}\left[\delta\left( \Q[f] - q\right)\middle|\bX_n, \by_n, \tilde\bx, \tilde y,\mathbf{S}_{n,m}, \mathbf{L}_{n,m}, \bpsi_m\right].
\end{equation}
The information gained through the hypothetical experiment $(\tilde{\bx}, \tilde{y})$ conditioned on the hyperparameters, say $G_{n,m}(\tilde{\bx}, \tilde{y})$ is given by the KLD between $p_{n,m}(q|\tilde{\bx}, \tilde{y})$ and $p_{n,m}(q)$.
Mathematically, it is:
\begin{equation}
    \label{eqn:kld}
    G_{n,m}(\tilde{\bx}, \tilde{y}) = \int_{-\infty}^{\infty}p_{n,m}(q|\tilde{\bx}, \tilde{y})\log\frac{p_{n,m}(q|\tilde{\bx}, \tilde{y})}{p_{n,m}(q)}dq.
\end{equation}
The expected information gain of the hypothetical experiment conditioned on the $m$-th posterior sample of the latent variables, say $G_{n,m}(\tilde{\bx})$, is obtained by taking the appropriate expectation of $G_{n,m}(\tilde{\bx},\tilde{y})$ over $\tilde{y}$.
Specifically,
\begin{equation}
    \label{eqn:ideal_G_nm}
    G_{n,m}(\tilde{\bx}) = \int_{-\infty}^{\infty} G_{n,m}(\tilde{\bx}, \tilde{y})p(\tilde{y}|\tilde{\bx},\bX_n,\by_n,\mathbf{L}_{n,m},\mathbf{S}_{n,m},\bpsi_{m},\sigma)d\tilde{y}.
\end{equation}
Finally, we construct the expected information of a hypothetical experiment by averaging over all the posterior samples of the latent variables, i.e.,
\begin{equation}
    \label{eqn:expected_info_gain}
    G_n(\tilde\bx) = \mathbb{E}\left[G_{n,m}(\tilde\bx)\middle|\bX_n,\by_n\right].
\end{equation}
The next experiment or simulation is selected by solving:
\begin{equation}
    \label{eqn:iaf_def_gen}
    \bx_{n+1} = \arg\max_{\tilde{\bx}} G_n(\tilde{\bx}).
\end{equation}

\subsection{Numerical approximation of the expected information gain}
\label{sec:current_state_of_knowledge}
Unfortunately, it is not possible to calculate \qref{expected_info_gain} analytically.
To overcome this difficulty, we derive a rather crude, but effective, Monte Carlo approximation.
We call this approximation $\hat{G}_n(\tilde\bx)$ and it is defined as follows:
\begin{equation}
    \label{eqn:G_approx}
    \hat{G}_n(\tilde\bx) := \frac{1}{M}\sum_{m=1}^M \hat{G}_{n,m}(\tilde \bx),
\end{equation}
where $m=1,\dots,M$ sums over posterior samples of hyper-parameters and latent variables,
\begin{equation}
    \hat{G}_{n,m}(\tilde\bx) := \frac{1}{B}\sum_{b=1}^B \hat{G}_{n,m}(\tilde\bx, \tilde{y}_b),
\end{equation}
where $b=1,\dots,B$ sums over samples of the response from the point-predictive distribution conditional on the $m$-th hyper-parameter and latent variable sample.
The term $\hat{G}_{n,m}(\tilde\bx,\tilde{y}_b)$ must approximate \qref{ideal_G_nm}.
It is not trivial to create a simple unbiased estimator of this term for an arbitrary non-linear quantity of interest, as the densities $p_{n,m}(q)$ and $p_{n,m}(q|\tilde\bx, \tilde y)$ are not analytically available.
To deal with this final barrier, we approximate these densities with Gaussians.
For example, the first one is approximated by:
\begin{equation}
    \label{eqn:dist_1}
    p_{n,m}(q)\approx\mathcal{N}\left(q\middle|\mu_{n,m,1}, \sigma_{n,m,1}^2\right),
\end{equation}
where $\mu_{n,m,1}$ and $\sigma_{n,m,1}^2$ are unbiased estimators of the mean and the variance, respectively, constructed by taking $S$ samples from $p_{n,m}(q)$.
Similarly, the second one is approximated by:
\begin{equation}
    p_{n,m}(q|\tilde\bx,\tilde{y})) = \mathcal{N}\left(q\middle|\mu_{n,m,2}(\tilde\bx,\tilde y), \sigma_{n,m,2}^2(\tilde\bx,\tilde y)\right).
\end{equation}
Now, using the known formula for the Kullback-Leibler divergence between two Gaussians, we set:
\begin{equation}
    \label{eqn:kld_Gaussian}
    \begin{array}{ccc}
    \hat{G}_{n,m}(\tilde\bx, \tilde{y}_b) &:=&\log\left(\frac{\sigma_{n,m,1}}{\sigma_{n,m,2}(\tilde{\bx}, \tilde{y}_b)}\right) + \frac{{\sigma^{2}_{n,m,2}(\tilde{\bx}, \tilde{y}_b)}}{2{\sigma^{2}_{n,m,1}}}  + \frac{{(\mu_{n,m,2}(\tilde{\bx},\tilde{y}_b) - \mu_{n,m,1})}^{2}}{2{\sigma^{2}_{n,m,1}}} - \frac{1}{2}.
    \end{array}
\end{equation}

\subsubsection{Maximizing the expected information gain}
The expected information gain is the expected Kullback-Leibler divergence (EKLD) from the a posteriori distribution of $Q$ to the a priori distribution of $Q$.
Maximizing the EKLD defined in \qref{iaf_def_gen} might ideally need a multi-start-optimization algorithm, but we resort to a Bayesian global optimization algorithm, based on augmented expected improvement~\cite{huang2006global}, (see Algorithm~\ref{alg:bgo_gen}), to maximize the EKLD.
In our experiments, we use $N_f=30$ BGO iterations to optimize the EKLD for one-dimensional functions. 
For multi-dimensional functions $N_f=40$ BGO iterations are used to optimize the EKLD.
\subsection{Selecting the number of initial data points}
The number of initial data points, denoted by $N_i$, is taken as low as possible to test the limits of the methodology.
In most literature, as a rule of thumb, $5d$ to $10d$ number of initial samples are used. 
Readers interested in the problem of the optimal selection of initial data size can refer to the work of S\`obester et al.~\cite{sobester2005design} where the authors discuss the problem in the context of optimization.
The problem of selecting an optimal number of initial points is beyond the scope of the work presented here.

\begin{algorithm}[htb]
\caption{Optimizing the EKLD using AEI-based BGO.}
\begin{algorithmic}[1]
\Require Number of initial data for EKLD GP $N_i$;
         Total number of EKLD evaluations $N_f$;
         Number of candidate designs $N_d$ for BGO;
         Number $M$ of HMC samples from the posterior of the hyperparameters;
         The index of maximum value of EKLD in the corresponding set, $k^{*}$.
         Stopping tolerance $\tau >0$.
    \State Evaluate $\hat{G}(\tilde{\bx})$ using \qref{G_approx} at the $N_i$ points selected using Latin Hypercube Sampling (LHS)~\cite{mckay2000} to obtain training data, $\tilde{\bX}_{N_i} = \left\{\tilde{\bx}_1,\dots,\tilde{\bx}_{N_i}\right\}$ and $\mathbf{G}_{N_i} = \left\{\tilde{G}_1 = G(\bx_1),\dots,\tilde{G}_{N_i} = G(\bx_{N_i})\right\}$, for BGO.
    \State $n \leftarrow N_i$.
            \While {$n < N_f$} 
                \State Fit a vanilla GP on the input-output pairs $\tilde{\bX}_{n}$-$\tilde{\mathbf{G}}_n$.
                \State Generate a set of candidate test points $\hat{\bX}_{N_d}=\left\{\hat{\bx}_1,\dots,\hat{\bx}_{N_d}\right\}$ using LHS.
                \State Compute the AEI for all of the candidate points in $\hat{\bX}_{N_d}$.
                \State Select ${\hat{\bx}}_j$ to be the point that exhibits the maximum AEI.
                \If{If the maximum AEI is smaller than $\tau$}
                    \State Break.
                \EndIf
                \State Obtain $\hat{G}_j = \hat{G}(\hat{\bx}_j)$ using \qref{G_approx}.
                \State $\tilde{x}_{n + 1} \leftarrow \hat{x}_j$.
                \State $\tilde{G}_{n + 1} \leftarrow \hat{G}_j$.
                \State $\bX_{n + 1} \leftarrow \tilde{\bX}_{n}\cup\{{\tilde{\bx}}_{n + 1}\}$.
                \State $\mathbf{G}_{n + 1} \leftarrow \mathbf{G}_{n}\cup\{\tilde{G}_{n + 1}\}$.
                \State $n\leftarrow n + 1$.
    \EndWhile
    \State $k^{*} \leftarrow {\arg\max}{\tilde{\mathbf{G}}}_{N_f}$
    \State return $\tilde{\bX}_{N_f, k^{*}}$.
\end{algorithmic}
\label{alg:bgo_gen}
\end{algorithm}

\begin{algorithm}[htb]
\caption{Bayesian optimal design of experiments maximizing the expected information gain a statistical QoI of a physical response.}
\begin{algorithmic}[1]
\Require Initially observed inputs $\bX_{N_i}$;
         initially observed outputs $\bY_{N_i}$;
         maximum number of allowed experiments $N_{f}$.
    \State $n \leftarrow N_i$.
            \While {$n < N_{f}$}
                \State Sample the latent function and hyper-parameters from \qref{post_gp_full}, to quantify apriori and a posteriori state of knowledge about the QoI.
                \State Find the next experiment $\bx_{n+1}$ using Algorithm \ref{alg:bgo_gen} to solve \qref{iaf_def_gen}.
                \State Evaluate the objective at $\bx_{n+1}$ measuring $y_{n+1} = f(\bx_{n+1})$.
                \State $\bX_{n+1} \leftarrow \bX_{n}\cup\{\bx_{n+1}\}$.
                \State $\bY_{n+1} \leftarrow \bY_{n}\cup\{y_{n+1}\}$.
                \State $n\leftarrow n + 1$.
        \EndWhile
\end{algorithmic}
\label{alg:ekld_gen}
\end{algorithm}

\section{Numerical examples}
\label{sec:results-bode-generic}
We present results for the methodology's performance on two one-dimensional mathematical 
functions, a three-dimensional problem, and a five-dimensional problem. 
The input domain for the first two synthetic problems is $[0, 1]$ whereas for the third synthetic problem the input domain is $[-2, 6]^{3}$. 
The input space for the five dimensional numerical example becomes the hyper-cube $[0, 1]^{5}$. 
The number of initial data points is denoted by $n_{i}$. 

\subsection{Selecting hyperpriors}
\label{sec:hyperpriors-bode-generic}
The prior distributions for the one-dimensional functions are chosen as follows: a) the mean function on the log-lengthscale GP is fixed to a negative integer constant. 
The reason behind it is that we wish to encode prior information about the lengthscale taking values as low as of the order of 1e-1. 
Thus, defining a lower bound or threshold on the point estimates of the lengthscale values.
In this work, this constant mean function is fixed at -2 for one-dimensional problems. 
For higher dimensional problems, this constant is fixed to 0.
b) for the signal-variance we choose a Gaussian prior, $\mathcal{N}(0, 4)$, on the mean function of the log-signal-variance GP for the one-dimensional problems.
This is a relatively diffuse prior which complements the Bayesian approach.
c) for the multi-dimensional problems we fixed the mean function to a value of 0.

Another technique to choose these prior distributions could be the Bayesian information criterion (BIC)~\cite{schwarz1978estimating}. 
Under the BIC combinations of prior distributions on the hyperparameters are compared against each other based on value of the data likelihood and a penalty criterion which is a function of the number of data and parameters.
Since, the number of data and parameters would be the same, the BIC would boil down to maximizing the likelihood.
We look to choose the priors based on some basic intuition about GPs and some prior knowledge about the function.
The same prior distributions are used for all one-dimensional functions.
We do this because we wish to test a set of non-informative priors that can be used across different problems without the need for any user intervention.
The number of HMC samples for each problem is fixed at 11,500, from which the first 1500 samples are discarded.
For further details on the HMC part of training the NSGP, we refer the readers to~\cite{neal2011mcmc,gelman2013bayesian}.

We mentioned in the previous section that the values of the number of posterior samples of $\btheta_{m}$, denoted by $S$, number of samples of $\tilde{y}$ denoted by $B$ and the number of samples of the $Q$ denoted by $M$ are fixed at 50.
This is done to ensure a default setting for the different controls of the algorithm irrespective of the function or the QoI being inferred.
In some cases, for some QoIs like the estimating the 2.5th percentile, or inferring the minimum or maximum values of a multi-modal function the default settings are changed to obtain smooth convergence results which can be explained better.
However, we do ensure that the settings remain the same for the ELKD and the other methods in all of our comparison studies.

\subsection{Numerical example 1}
\label{sec:toy_1_gen}
Consider the following function:
\begin{equation}
    f(x) = 4\left( {1 - \sin\left(6x + 8{e^{6x - 7}}\right)}\right),
    \label{eqn:toy_1_gen}
\end{equation}
defined on $x \in [0,1]$.
This function is smooth throughout its domain, but it exhibits two local minima.
We will apply our methodology to estimate the QoIs in~\sref{metho-bode-generic} including the case of inferring the 2.5th percentile of the function.
The true values of the five $\Q[f]$s enumerated in~\sref{metho-bode-generic} are:
\begin{enumerate}
    \item $\mathbb{E}[f] = -1.36$
    \item $\mathbb{V}[f] = 0.30$
    \item $\min [f] = -2.00$
    \item $\mathbb{Q}_{2.5}[f]= -1.99$
\end{enumerate}

We apply our methodology to this problem starting from $n_i=3$ and sample a total of $N=18$ points.
\begin{figure}[!htbp]
\centering
    \subfigure[]{
        \includegraphics[width=.45\columnwidth]{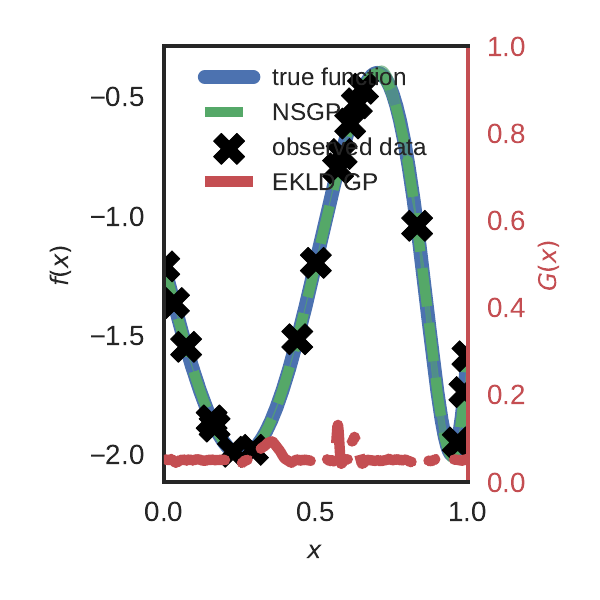}
    }
    \subfigure[]{
        \includegraphics[width=.45\columnwidth]{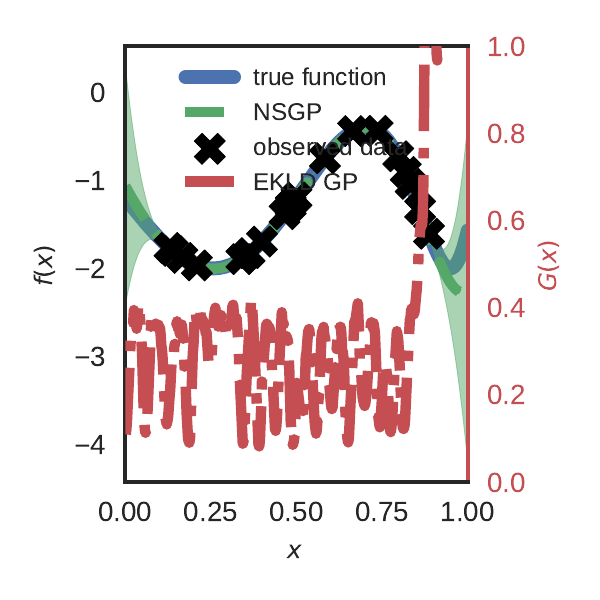}
    }
    \subfigure[]{
        \includegraphics[width=.45\columnwidth]{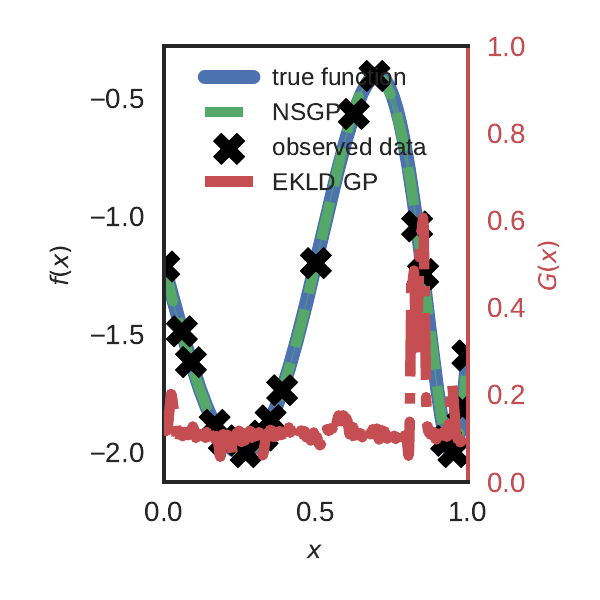}
    }
    \subfigure[]{
        \includegraphics[width=.45\columnwidth]{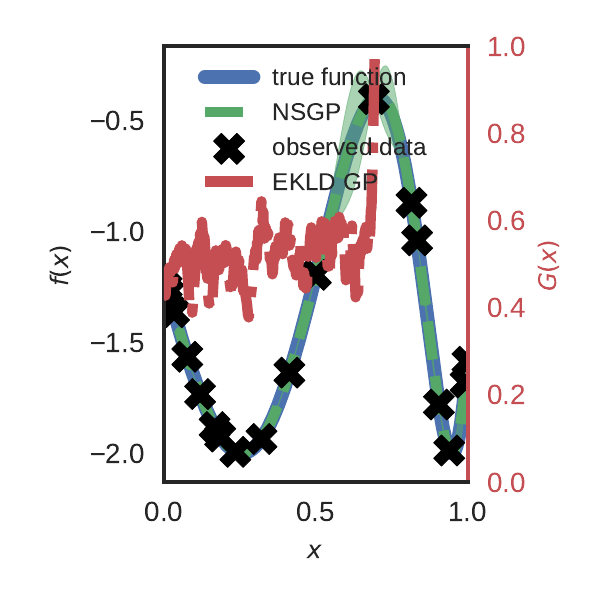}
    }
    \caption{One-dimensional synthetic problem ($n_{i}=3$) shows the state of the function at the end (15th iteration) of the algorithm for inferring:
        Subfigures ~(a) $\mathbb{E}[f]$,
        ~(b) $\mathbb{V}[f]$,~(c)
        $\min [f]$, and ~(d) $\mathbb{Q}_{2.5}[f]$.
    }
    \label{fig:toy_1_state}
\end{figure}

\fref{toy_1_state} shows the final state of designs for the different $\Q[f]$s.
The thick blue line represents the true function $f$, \qref{toy_1_gen}.
The black crosses are the observed data at the final stage.
In Subfigure~(a), the next experiment selected by maximizing the EKLD, see Algorithm \ref{alg:bgo_gen}, corresponds to the purple diamond.
The mean of the GP fit to the expected information gain $G(\tilde{\bx})$ constructed by BGO in Algorithm~\ref{alg:bgo_gen}.
The predictive mean of the EKLD is shown by the dotted light blue line.
This dotted line represents the response surface of the EKLD after the BGO has ended and the red shaded area around it represents the uncertainty (2.5 percentile and 97.5 percentile) around the mean.
As expected, the mean of the EKLD is very small or close to zero at points where experiments have been performed.
Thus, the point selected by the methodology (purple diamond) is located in the input space where the EKLD has high mean.
The posterior mean of the GP of the black-box function is represented by the dashed bottle-green line.
The bottle-green shaded area represents the uncertainty (2.5 percentile and  97.5 percentile) around it.

\subsubsection{Inferring lengthscale and signal-variance}
The inferred lengthscale and signal-variance GPs are shown in \fref{toy_1_stats}~(b) and~(c) respectively for the case of inferring the statistical expectation.
These plots simply show the posterior mean using each of the S posterior samples.
The lengthscale is larger in the region $[0, 0.6]$ compared to $[0.6, 1]$.
This can be understood by comparing the waviness of the function in these regions.
Similarly, the inferred signal-variance has higher absolute value corresponding to those taken by $f$.
With limited data, fully-Bayesian HMC allows one to obtain such estimates of uncertainty around the inferred value of lengthscale and signal-variance.
Other approaches which come at a lower computational cost, like maximum a posteriori (MAP) or maximum likelihood estimate (MLE) would need a significantly larger number of training points to infer meaningful point estimates for the lengthscale and the signal-variance.
In the low-sample regime the MAP estimate for the hyperparameters are prone to mislead the methodology either by selecting a single sample i.e. a local optima of the posterior of the hyperparameters.
Multiple optimization restarts for MAP and MLE approaches might be beneficial but only slightly unless the number of restarts is of the order of 100.
This usually increases the computational burden without making the solution significantly better.
The fully-Bayesian approach used here makes a compelling case to infer the lengthscale and signal-variance GPs under epistemic uncertainty albeit at a higher computational cost.

\begin{figure}[!htbp]
\centering
    \subfigure[]{
        \includegraphics[width=.45\columnwidth]{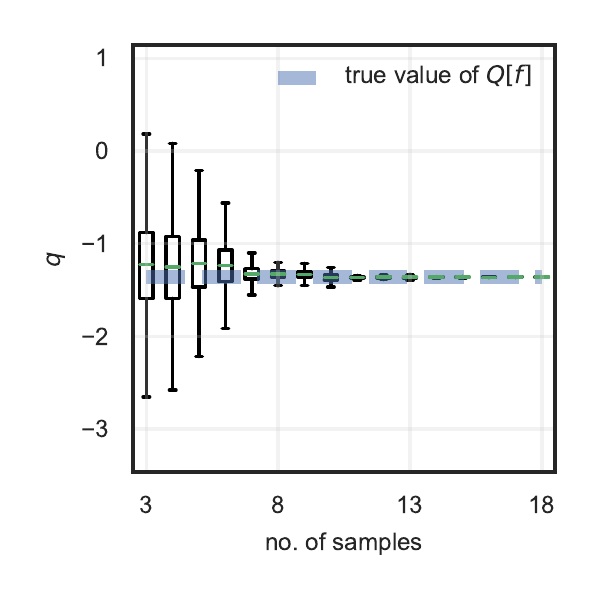}
    }
    \subfigure[]{
        \includegraphics[width=.45\columnwidth]{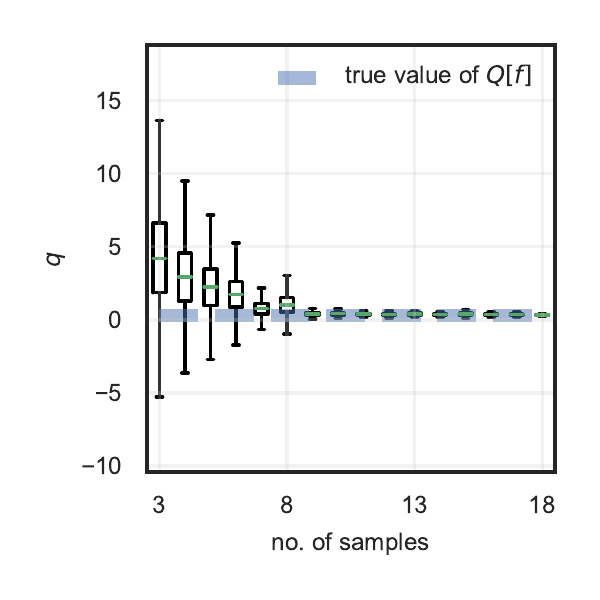}
    }
    \subfigure[]{
        \includegraphics[width=.45\columnwidth]{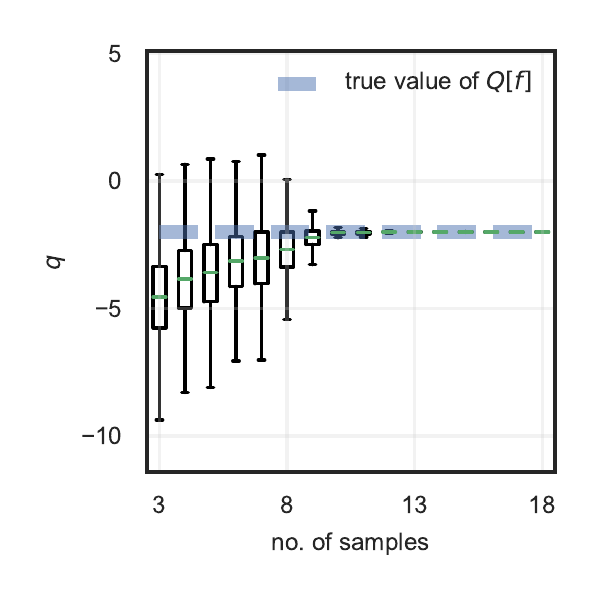}
    }
    \subfigure[]{
        \includegraphics[width=.45\columnwidth]{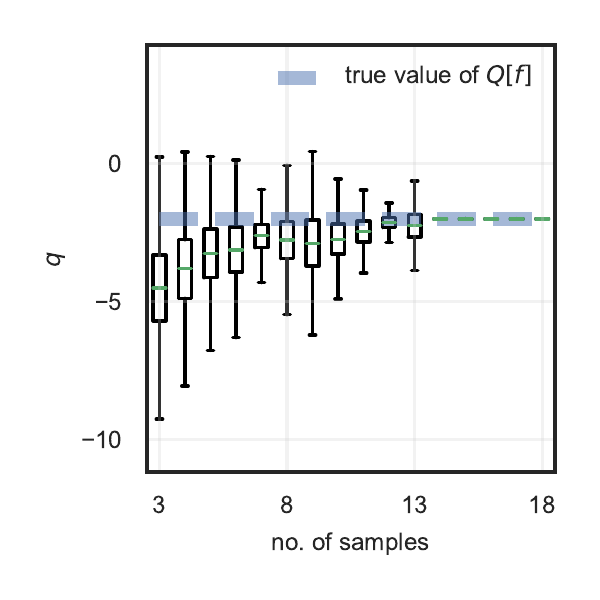}
    }
    \caption{One-dimensional synthetic problem ($n_{i}=3$) shows the convergence of EKLD to the true value of $Q$ for inferring:
    Subfigures ~(a) $\E[f]$,
    ~(b) $\V[f]$,~(c)
    $\min [f]$, and ~(d) $\mathbb{Q}_{2.5}[f]$.
    }
    \label{fig:toy_1_box}
\end{figure}

\begin{figure}[!htbp]
\centering
    \subfigure[]{
        \includegraphics[width=.45\columnwidth]{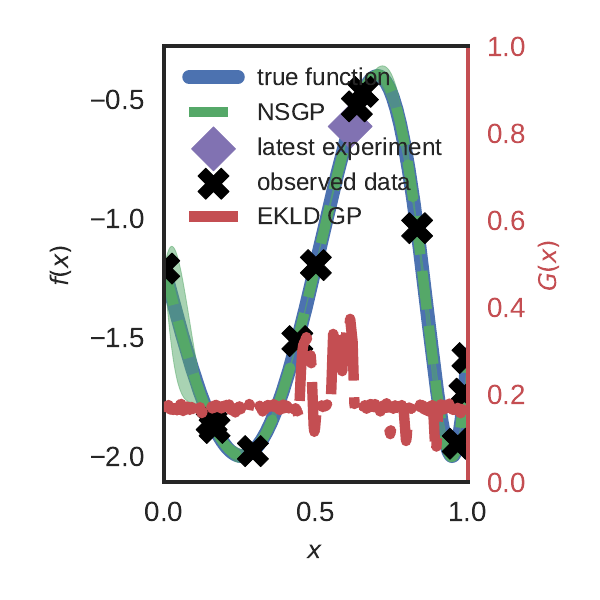}
    }
    \subfigure[]{
        \includegraphics[width=.45\columnwidth]{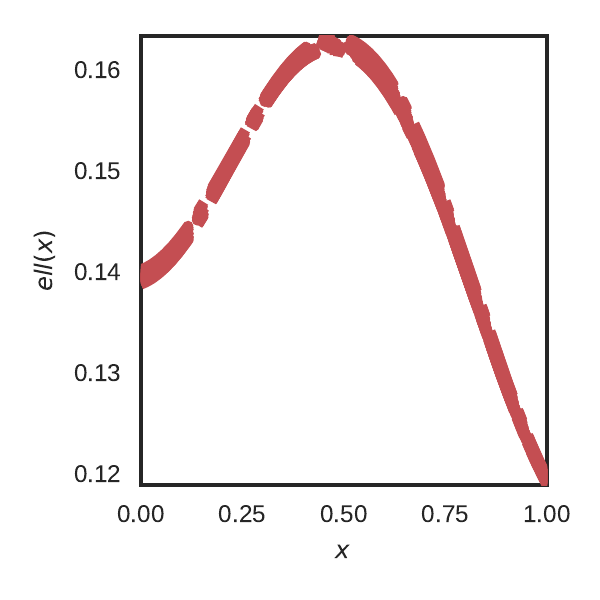}
    }
    \subfigure[]{
        \includegraphics[width=.45\columnwidth]{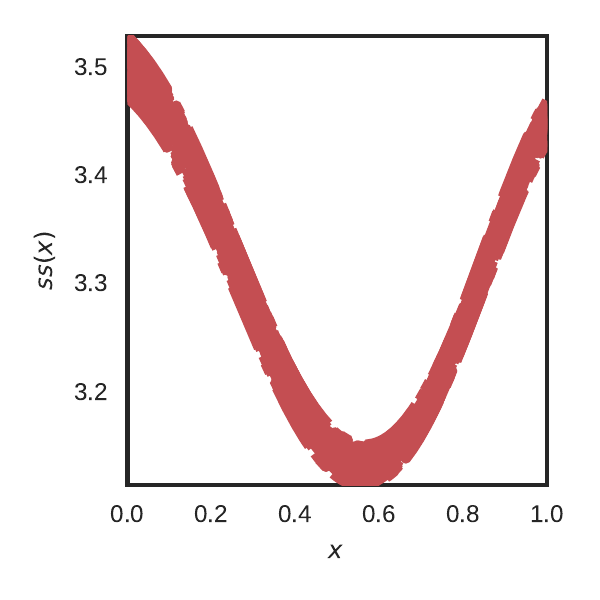}
    }
    \caption{One-dimensional synthetic problem ($n_{i}=3$) shows the statistics of the
    FBNSGP at the 10th iteration of the sampling where:
    Subfigure ~(a) shows the state of the sampling,
    ~(b) shows the inferred point estimates of the lengthscale 
    and~(c) shows the inferred point estimates of the signal-variance.
    }
    \label{fig:toy_1_stats}
\end{figure}

\subsection{Numerical example 2}
\label{sec:toy_2_gen}
We consider the following Gaussian mixture function to test and validate our methodology further.
\begin{equation}
    \label{eqn:toy_2_gen}
    \begin{array}{ccc}
    f(x) &=&  \frac{1}{\sqrt{2\pi}s_{1}}\exp\left\{-\frac{(x-m_{1})^{2}}{2{s_{1}}^2}\right\}\\
    && + \frac{1}{\sqrt{2\pi}s_{2}}\exp\left\{-\frac{(x-m_{2})^{2}}{2{s_{2}}^2}\right\},
    \end{array}
\end{equation}
where $m_{1}=0.2$ and $s_{1}=0.05$, $m_{2}=0.8$ and $s_{2}=0.05$.
As we can see from \qref{toy_2_gen}, the function is a sum of probability densities of two Gaussian distributions. 
The function has two narrow areas of high magnitude.
The true value of the $\Q[f]$s are analytically available, and take the following values:
\begin{enumerate}
    \item $\mathbb{E}[f] = 2.00$
    \item $\mathbb{V}[f] = 7.28$
    \item $\min [f] = 0.00$
    \item $\mathbb{Q}_{2.5}[f]= 0.00$
\end{enumerate}

\subsubsection{Inferring lengthscale and signal-variance}
We use the same hyperpriors for this problem as in \sref{toy_1_gen}. 
The inferred signal-variance and lengthscale can be seen in \fref{toy_2_stats} for iteration no. 22 of sampling. 
The lengthscale values in \fref{toy_2_stats}~(b) show high values in the middle region of the input space and lower values in the areas where the input value is 0.2 and 0.8 respectively.
This behavior of the inferred lengthscale GP is in concurrence with the true function being inferred in \fref{toy_2_stats}~(a) where the methodology has almost learned the true function.
The lengthscale values should be small as the waviness is high in areas of the two sharp peaks in the function. 
Similarly, the lengthscale values should be high where the function is very smooth or in other words flat.
The inferred signal-variance, shown in \fref{toy_2_stats}~(c), also corresponds to the scalar value of the true function in the corresponding regions of the input space.

\subsubsection{Understanding convergence plots}
For this problem, the methodology starts with $n_i=5$ and samples another 25 points.
The final state of sampling for each $\Q[f]$ can be seen in \fref{toy_2_state}.
The final states show the different sets of designs obtained for different QoIs.
The convergence of the estimated mean to the true value for each $\Q[f]$ and the reduction in uncertainty around the $\Q[f]$ can be seen in \fref{toy_2_box}.

A possible misinterpretation of convergence can arise if one considers the rise in uncertainty around the mean of $Q$ during the initial stages of sampling to be an anomaly in the computation of the information gain.
An example of this can be seen in \fref{toy_2_box} (b) as the uncertainty around the $Q$ increases as the number of samples increases from 5 to 6.
This is mainly due to stark difference in the values of hyperparameters between the two stages of sampling.
Thus, giving rise to different values for the inferred signal-variance or lengthscale.
We show this phenomenon for this case in \fref{toy_2_mcmc} which shows how the inferred signal-variance changed scale after the 9.
Specifically, in this case it happened because the methodology discovered a point in the bump which has a significantly higher value of $f$ as seen in \fref{toy_2_mcmc}~(c) and~(d).
This phenomenon is expected in the initial stages of the sampling or the low-sample regime since the methodology is still discovering the different areas of the response surface some of which may have veritably different signal-variance or lengthscale values.
Another reason that may lead to a sudden increase in uncertainty is the multi-modal nature of the posterior of the hyperparameters.
Although, the use of HMC enables the methodology to generate samples from the multi-modal posterior, there is always a possibility that the samples used in computing the information gain do not cover all the modes or worse, come from a single mode.
This is a well-known phenomenon with MCMC.

However, once a significant number of data have been collected the methodology rarely shows such behaviour.
This is corroborated by the results presented in this paper.
\begin{figure}[!htbp]
\centering
    \subfigure[]{
        \includegraphics[width=.45\columnwidth]{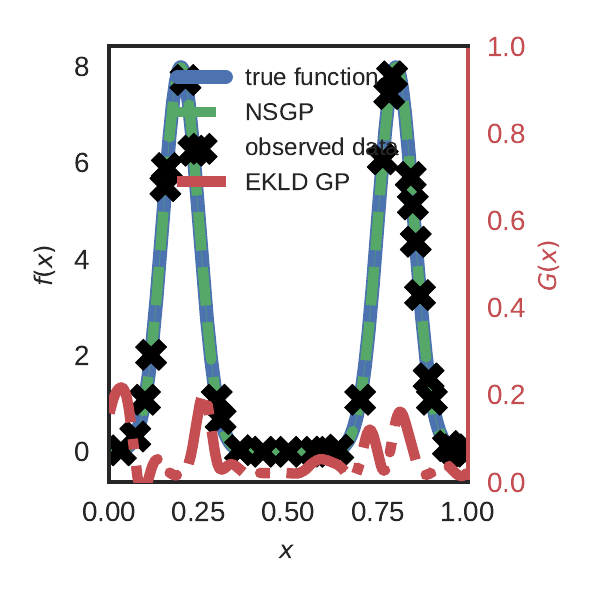}
    }
    \subfigure[]{
        \includegraphics[width=.45\columnwidth]{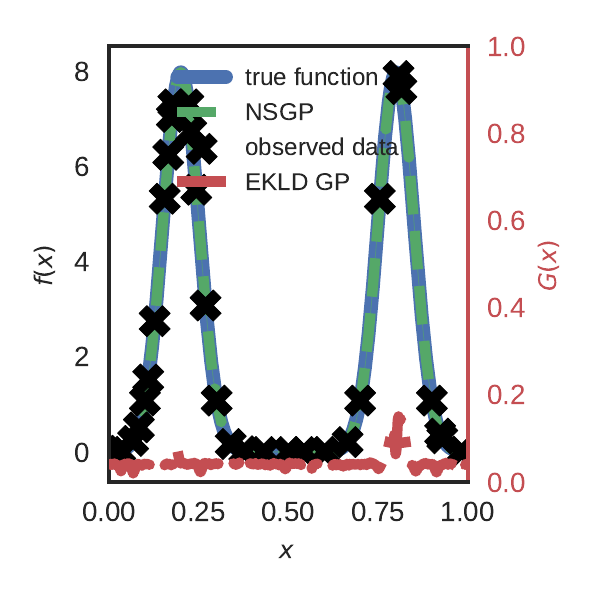}
    }
    \subfigure[]{
        \includegraphics[width=.45\columnwidth]{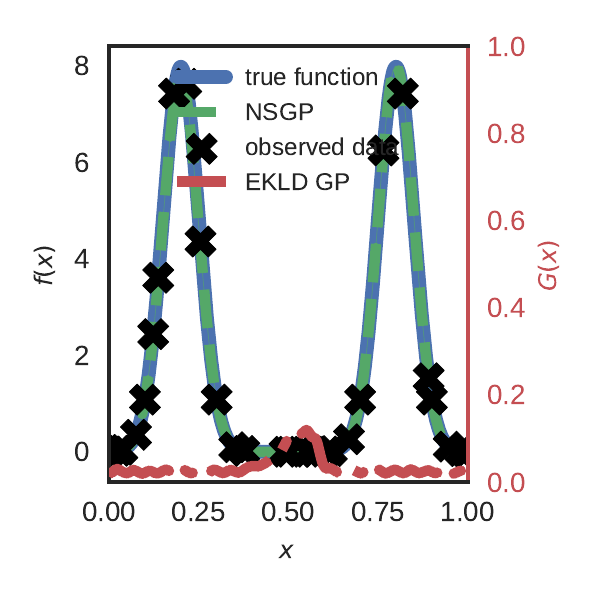}
    }
    \subfigure[]{
        \includegraphics[width=.45\columnwidth]{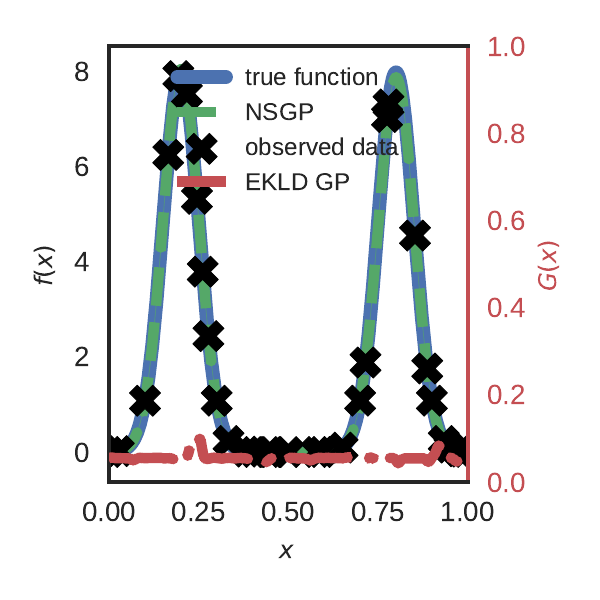}
    }
    \caption{One-dimensional synthetic problem ($n_{i}=5$) shows the state of the function at the end (25th iteration) of the algorithm for inferring:
        Subfigures ~(a) $\mathbb{E}[f]$,
        ~(b) $\mathbb{V}[f]$,~(c)
        $\min [f]$, and ~(d) $\mathbb{Q}_{2.5}[f]$.
    }
    \label{fig:toy_2_state}
\end{figure}
\begin{figure}[!htbp]
\centering
    \subfigure[]{
        \includegraphics[width=.45\columnwidth]{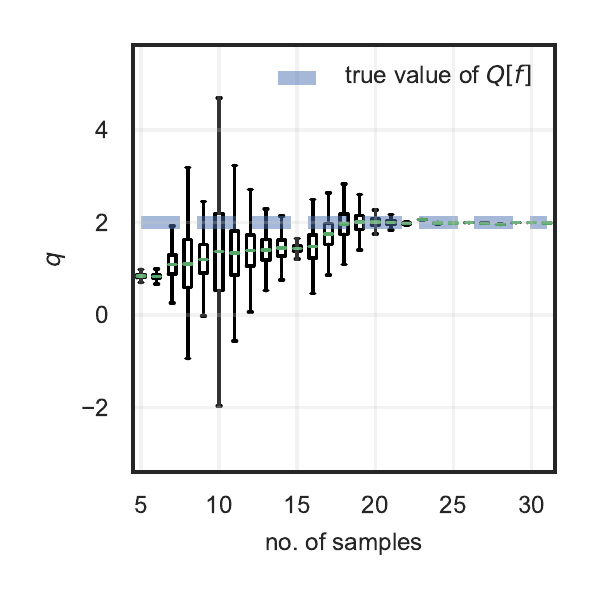}
    }
    \subfigure[]{
        \includegraphics[width=.45\columnwidth]{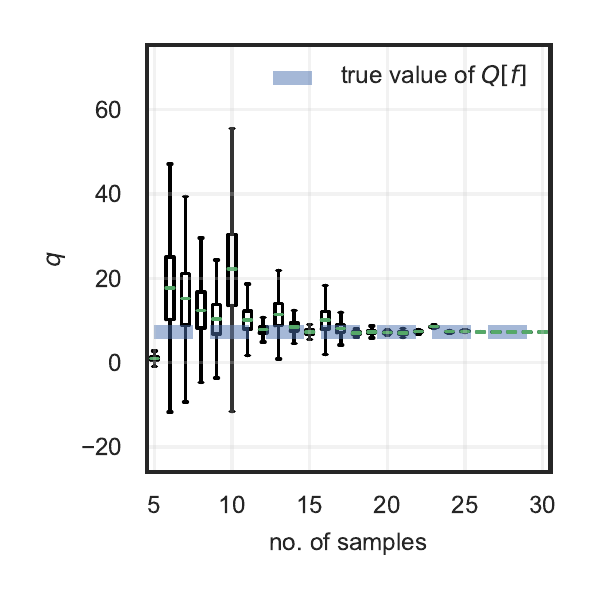}
    }
    \subfigure[]{
        \includegraphics[width=.45\columnwidth]{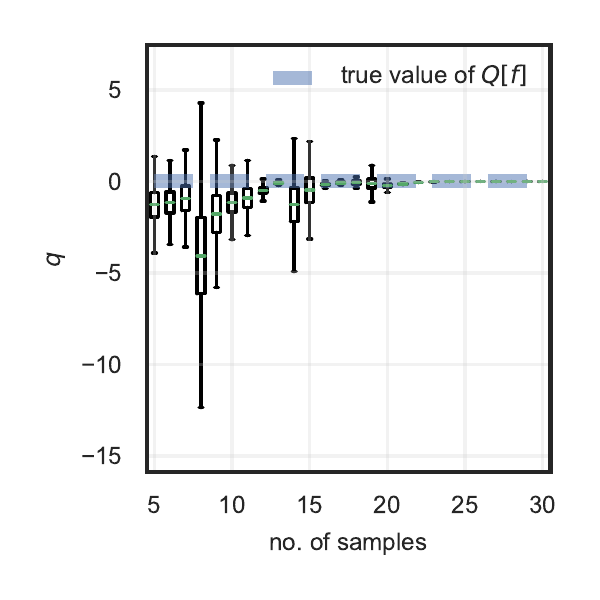}
    }
    \subfigure[]{
        \includegraphics[width=.45\columnwidth]{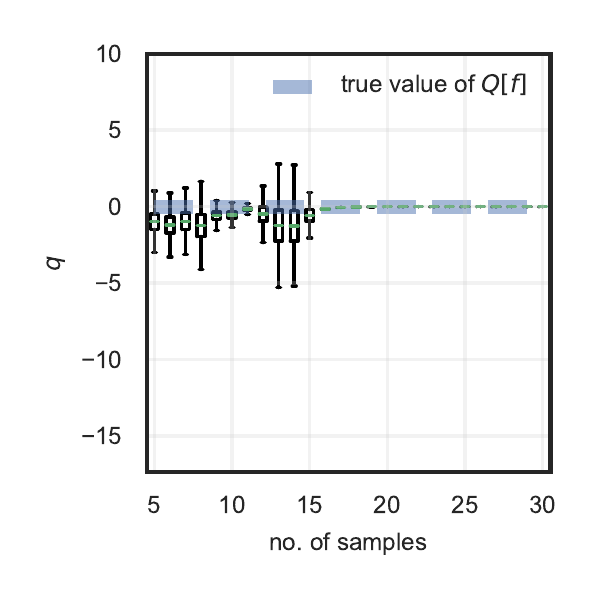}
    }
    \caption{One-dimensional synthetic problem ($n_{i}=5$) shows the convergence of EKLD to the true value of $Q$ for inferring:
    Subfigures ~(a) $\E[f]$,
    ~(b) $\V[f]$,~(c)
    $\min [f]$, and ~(d) $\mathbb{Q}_{2.5}[f]$.
    }
    \label{fig:toy_2_box}
\end{figure}

\begin{figure}[!htbp]
\centering
    \subfigure[]{
        \includegraphics[width=.45\columnwidth]{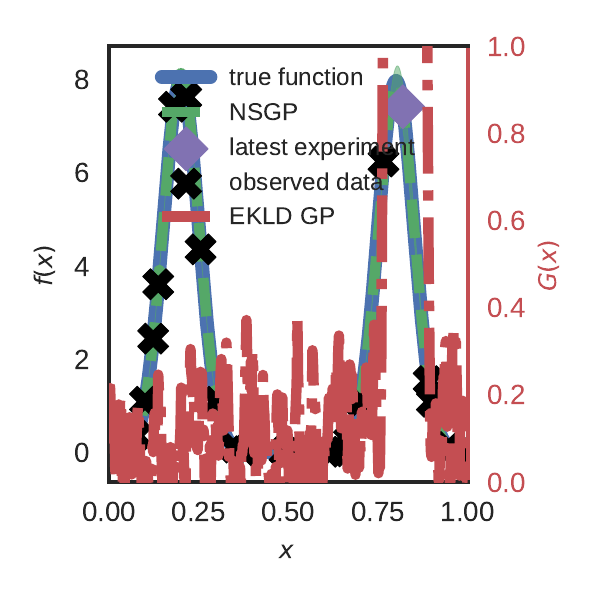}
    }
    \subfigure[]{
        \includegraphics[width=.45\columnwidth]{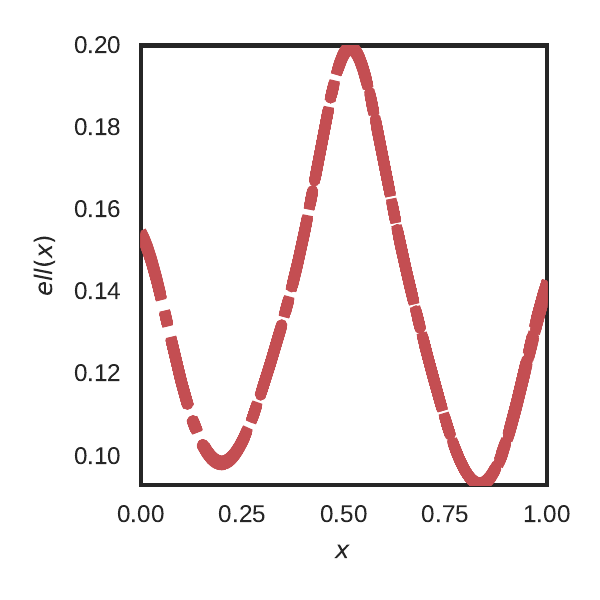}
    }
    \subfigure[]{
        \includegraphics[width=.45\columnwidth]{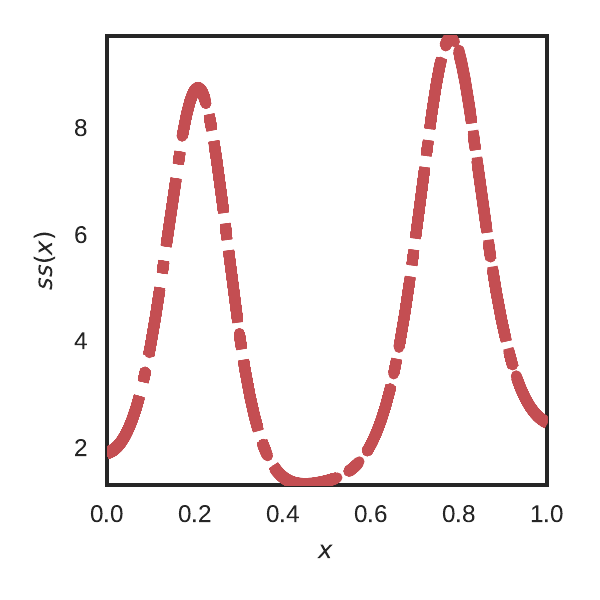}
    }
    \caption{One-dimensional synthetic problem ($n_{i}=5$) shows the statistics of the
    FBNSGP at the 23rd iteration of the sampling where:
    Subfigure ~(a) shows the state of the sampling,
    ~(b) shows the inferred point estimates of the lengthscale 
    and~(c) shows the inferred point estimates of the signal-variance.
    }
    \label{fig:toy_2_stats}
\end{figure}

\begin{figure}[!htbp]
\centering
    \subfigure[]{
        \includegraphics[width=.45\columnwidth]{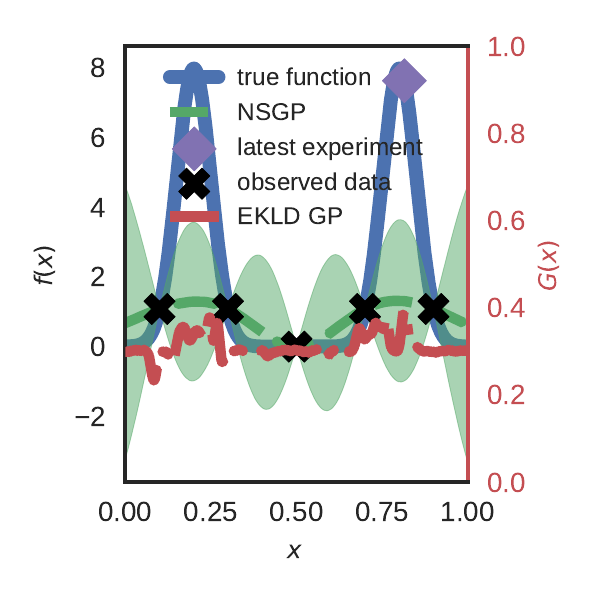}
    }
    \subfigure[]{
        \includegraphics[width=.45\columnwidth]{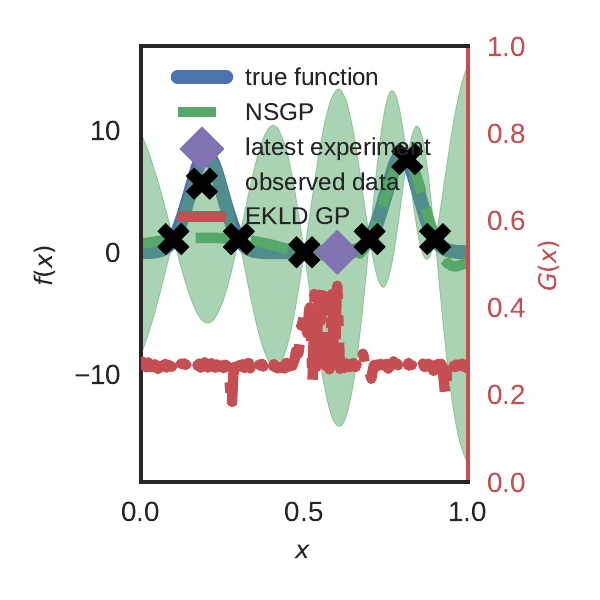}
    }
    \subfigure[]{
        \includegraphics[width=.45\columnwidth]{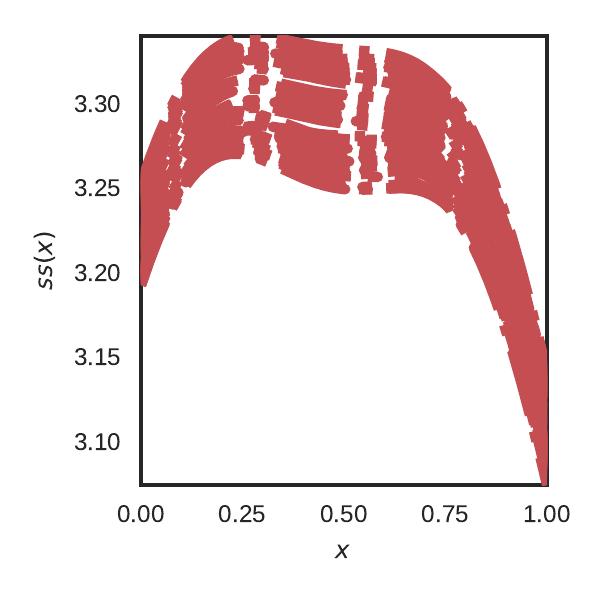}
    }
    \subfigure[]{
        \includegraphics[width=.45\columnwidth]{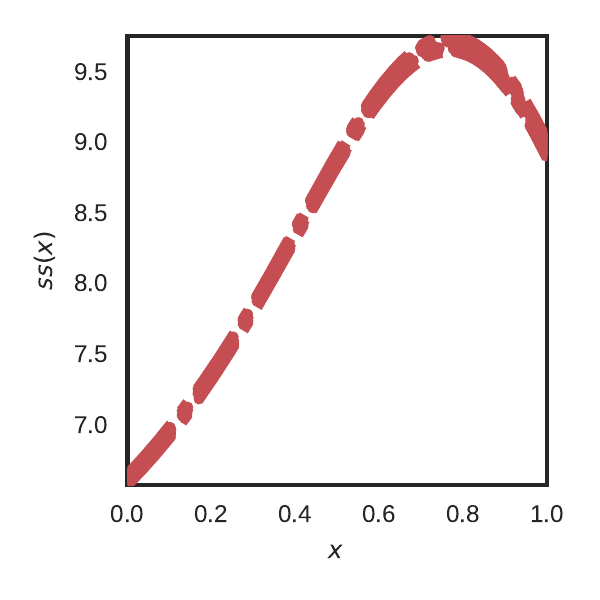}
    }
    \caption{One-dimensional synthetic problem ($n_{i}=5$) shows the statistics of the
    FBNSGP where:
    Subfigures ~(a) and~(b) show the state of the sampling, 
    and
    ~(c) and~(d) show the inferred point estimates of the signal-variance 
    at the end of the 1st and the 2nd iteration respectively.
    }
    \label{fig:toy_2_mcmc}
\end{figure}

\subsection{Numerical example 3}
\label{sec:toy_3_gen}
Consider the following three-dimensional function from \cite{dette2010generalized} to test our methodology further. 
\begin{eqnarray}
    f(\bx)&=&4(x_{1} + 8x_{2} - 8x_{2}^{2} - 2)^{2} + (3-4x_{2})^{2} \nonumber \\
    &&+ 16\sqrt{x_{3}+1}(2x_{3}-1)^{2}.
\label{eqn:toy_3_gen}
\end{eqnarray}
One difference between this function \qref{toy_3_gen} and the first two synthetic examples is the dimensionality of the problem.
This is crucial because the FBNSGP modeling framework is expected to behave in a slightly different manner in multiple input dimensions.
Unlike the one-dimensional numerical examples discussed above, we proceed with a constant zero mean function of the GPs that model the logarithms of the lengthscale and the signal-variance.
We also find this to be consistent with the BIC model selection at the beginning of the SDOE.
An intuitive explanation about the change in behaviour of lengthscale values in higher dimensions compared to lower dimensions is given in~\cite{aggarwal2001surprising}.
The true values of the $\Q[f]$s, analytically available, are:
\begin{enumerate}
    \item $\mathbb{E}[f] = -0.7864$
    \item $\mathbb{V}[f] = 0.0209$
    \item $\min [f] = -0.9999 $
    \item $\mathbb{Q}_{2.5}[f] = -0.9899 $
\end{enumerate}

We apply our methodology to this problem starting from $n_i=10$ and sample another 40 points.
\fref{toy_3_mean}~(b) shows that the methodology started with a highly uncertain estimate of the true value and eventually converged to a sharp peaked Gaussian distribution around the true value. 
The approximation to each $\Q[f]$ at each stage of the algorithm is shown in \fref{toy_3_mean}. 
The gradual reduction in uncertainty around each $\Q[f]$ also can be seen in \fref{toy_3_mean}. 

\begin{figure}[!htbp]
\centering
    \subfigure[]{
        \includegraphics[width=.45\columnwidth]{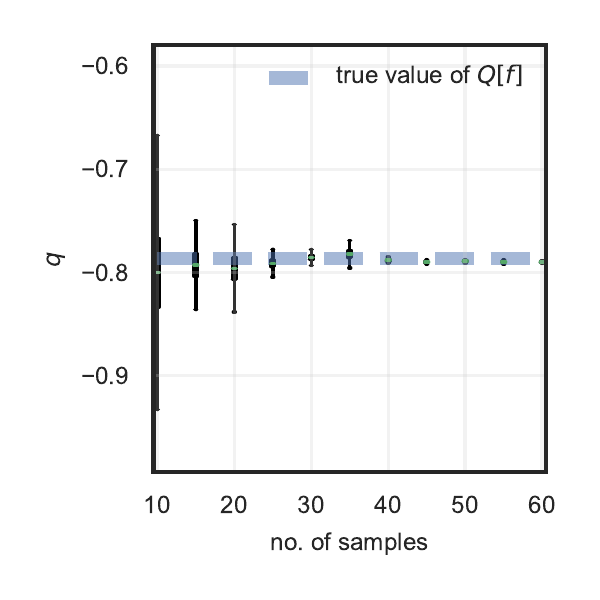}
    }
    \subfigure[]{
        \includegraphics[width=.45\columnwidth]{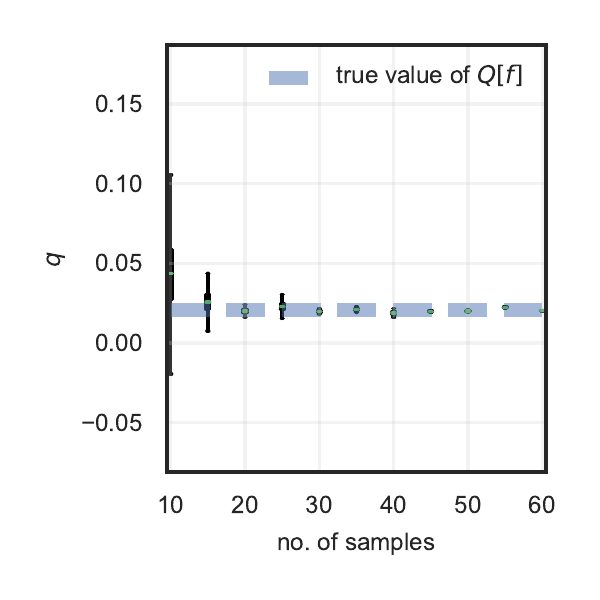}
    }
    \subfigure[]{
        \includegraphics[width=.45\columnwidth]{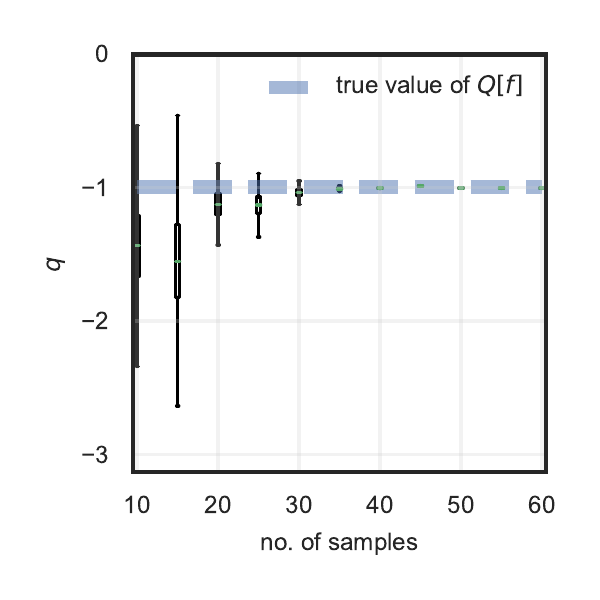}
    }
    \subfigure[]{
        \includegraphics[width=.45\columnwidth]{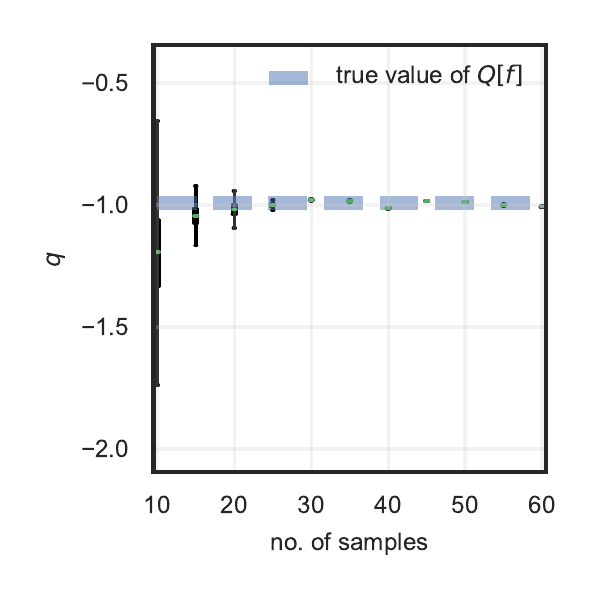}
    }
    \caption{Three-dimensional synthetic problem ($n_{i}=10$) shows 
    the convergence of EKLD to the true value of $Q$ for inferring:
    Subfigures ~(a) $\E[f]$,~(b) $\V[f]$,~(c)
    $\min [f]$, and ~(d) $\mathbb{Q}_{2.5}[f]$.
    }
    \label{fig:toy_3_mean}
\end{figure}

\subsection{Numerical example 4}
\label{sec:toy_4_gen}
The following five-dimensional function is taken from \cite{knowles2006parego}. 
\begin{eqnarray}
    f(\bx)&=& 10\sin(\pi x_{1}x_{2}) + 20 (x_{3}-5)^{2} +10x_{4} + 5x_{5}. 
\label{eqn:toy_4_gen}
\end{eqnarray}
This function \qref{toy_4_gen} is reasonably high-dimensional and challenging due to the non-linear input-output relation. 
The true values of the $\Q[f]$s, analytically available, are:
\begin{enumerate}
    \item $\mathbb{E}[f] = 0.3882$
    \item $\mathbb{V}[f] = 1.0896$
    \item $\min [f] = -1.5906 $
    \item $\mathbb{Q}_{2.5}[f] = -1.2782 $
\end{enumerate}
We apply our methodology to this problem starting from $n_i=10$ and sample another 60 points for inferring $\E[f]$ and $\V[f]$.
For inferring the 2.5th percentile~\fref{toy_4_mean}(e) of $f$ we start with 10 initial points and collect another 60 samples.
In the cases shown in \fref{toy_4_mean} (c) for inferring $\min[f]$y, the methodology starts with 20 initial points and samples another 50 points using the EKLD.
\begin{figure}[!htbp]
\centering
    \subfigure[]{
        \includegraphics[width=.45\columnwidth]{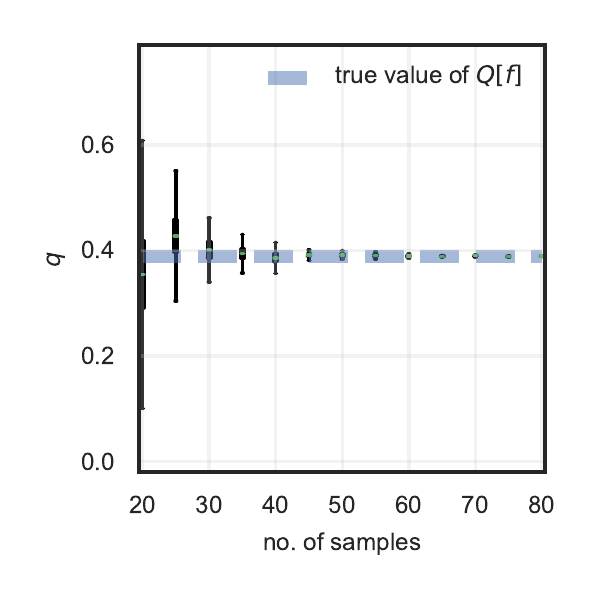}
    }
    \subfigure[]{
        \includegraphics[width=.45\columnwidth]{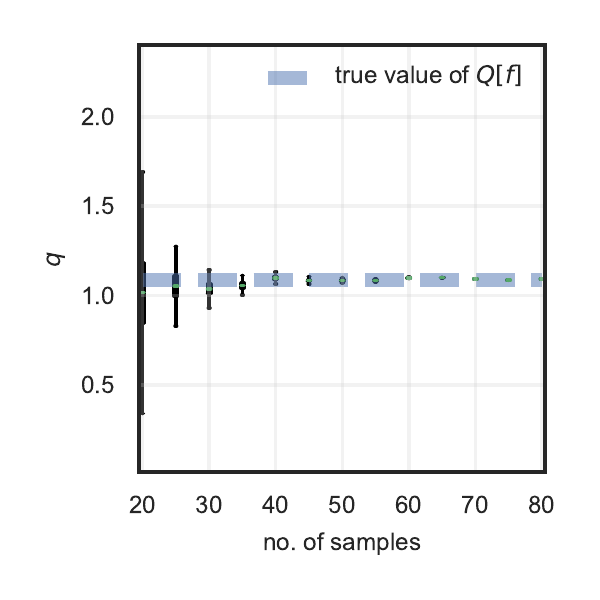}
    }
    \subfigure[]{
        \includegraphics[width=.45\columnwidth]{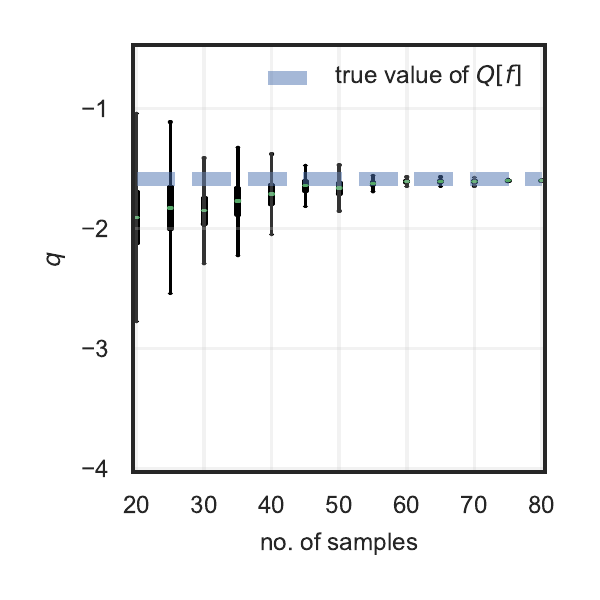}
    }
    \subfigure[]{
        \includegraphics[width=.45\columnwidth]{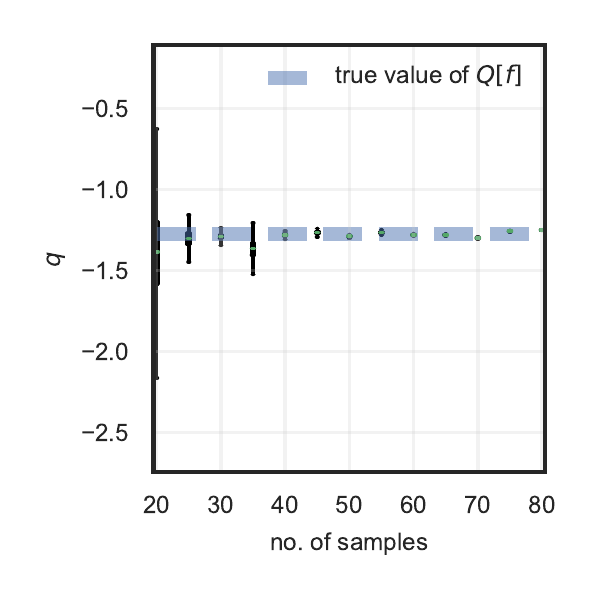}
    }
    \caption{Five-dimensional synthetic problem ($n_{i}=10$) shows 
    the convergence of EKLD to the true value of $Q$ for inferring:
    Subfigures ~(a) $\E[f]$, ~(b) $\V[f]$,~(c)
    $\min [f]$, and ~(d) $\mathbb{Q}_{2.5}[f]$.
    }
    \label{fig:toy_4_mean}
\end{figure}
The iteration-wise convergence of the $\Q[f]$s to the respective true value is shown in \fref{toy_4_mean}.
Along expected lines, as more samples are collected by the EKLD, the uncertainty around the mean of $\Q[f]$ reduces.
This uncertainty becomes negligible around the 50th sample mark for each of the five $\Q[f]$s in~\fref{toy_4_mean}.

\subsection{Wire drawing problem}
\label{sec:wire_gen}
The wire drawing process aims to achieve a required reduction in the cross section of the incoming wire, while aiming to infer statistics of the mechanical properties of the outgoing wire. 
The incoming wire passes through a series of dies (5 dies) to achieve an overall reduction in wire diameter.
An expensive FORTRAN code models the wire drawing process by modeling the micro-structure of the wire at each stage of the process using the Finite element method (FEM). 
We aim to infer statistics of the frictional work per Tonne (FWT) of the process.
The FWT is one of the outputs of the expensive code.
Rest of the technical regarding the micro-structure modeling using FEM remain the same as in Section 3.5 of~\cite{pandita2019bayesian}.

The true values of the $\Q[f]$s, analytically available, are:
\begin{enumerate}
    \item $\mathbb{E}[f] = -2.2402$
    \item $\mathbb{V}[f] = 0.1805$
    \item $\min [f] = -3.5724 $
    \item $\mathbb{Q}_{2.5}[f] = -3.027 $
\end{enumerate}
We apply our methodology to this problem starting from $n_i=10$ and 
sample another 70 points for inferring the different $\Q[f]$s.
\begin{figure}[!htbp]
\centering
    \subfigure[]{
        \includegraphics[width=.45\columnwidth]{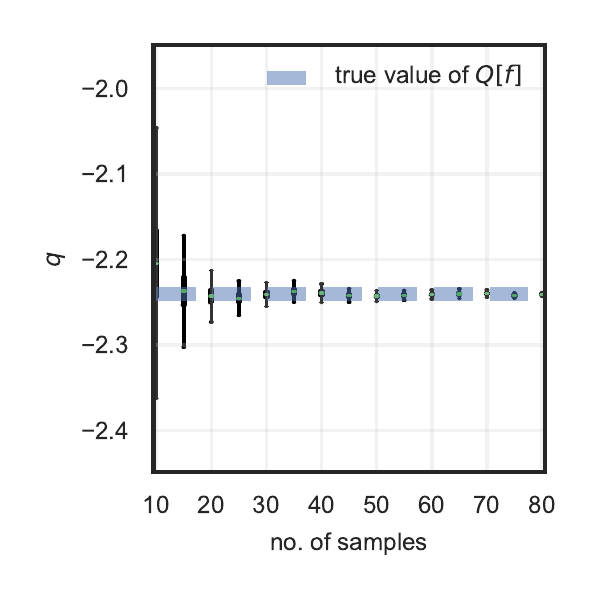}
    }
    \subfigure[]{
        \includegraphics[width=.45\columnwidth]{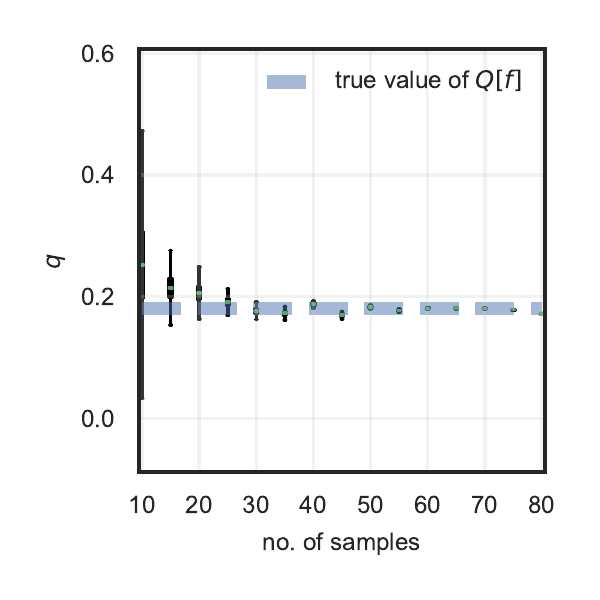}
    }
    \subfigure[]{
        \includegraphics[width=.45\columnwidth]{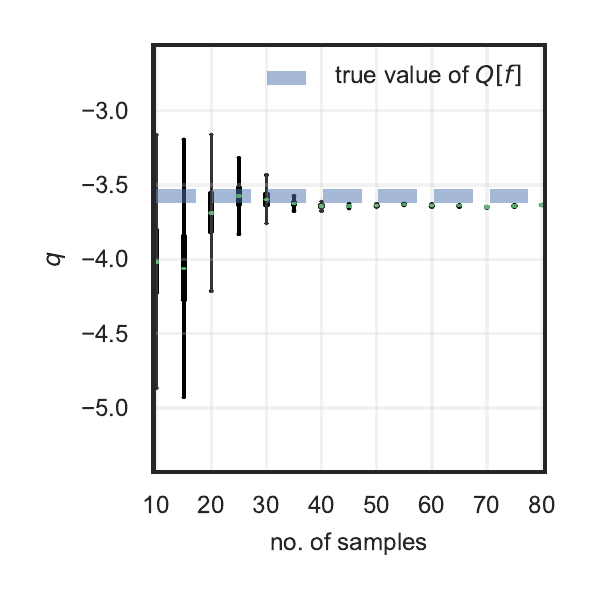}
    }
    \subfigure[]{
        \includegraphics[width=.45\columnwidth]{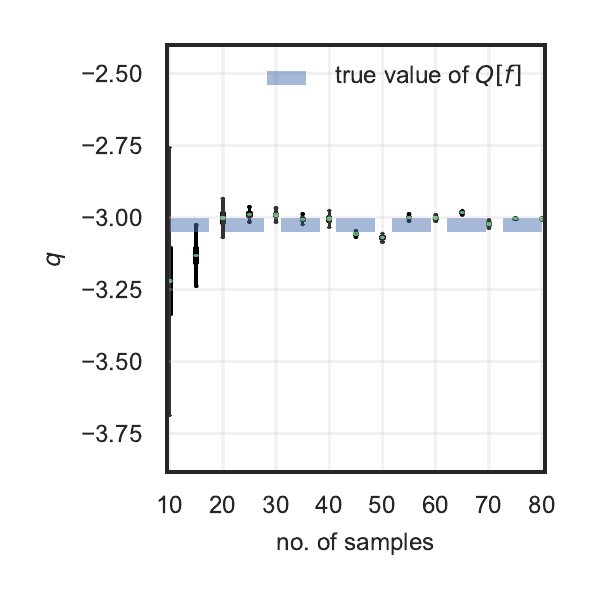}
    }
    \caption{Wire-drawing problem ($n_{i}=10$) shows 
    the convergence of EKLD to the true value of $Q$ for inferring:
    Subfigures ~(a) $\E[f]$,
    ~(b) $\V[f]$,~(c) 
    $\min [f]$, and ~(d) $\mathbb{Q}_{2.5}[f]$.
    }
    \label{fig:wire_mean}
\end{figure}

The iteration-wise convergence of the $\Q[f]$s to the respective true value is shown in \fref{wire_mean}.
It is interesting to note the noise in the convergence for $\Q[f]$ no.3 in \fref{wire_mean}~(c).
This is because the number of samples $M$ needed to approximate to $\Q[f]$s at each iteration for cases where a global minima is located in a small region becomes very high. 
One way around this could be to take more $M$ samples albeit at a very high computational cost.
For the $\Q[f]$s no.1, no.2 and no.4 EKLD seems to converge as the number of samples reaches 30.
Along expected lines, as more samples are collected by the EKLD, the uncertainty around the mean of $\Q[f]$ reduces.
The uncertainty around the expected value of the $Q$ becomes negligible around the 30th sample mark for each of the five $\Q[f]$s in~\fref{wire_mean}.

\section{Comparison studies}
\label{sec:comparison-bode-generic}
We compare the EKLD to two classic SDOE methods, namely uncertainty sampling (US) and expected improvement (EI). 
This is done in order to ascertain the convergence pattern of the EKLD to some extent. 
A comparison with US is done when the $\Q[f]$ is $\E[f]$, $\V[f]$ or $\mathbb{Q}_{2.5}[f]$.
This is because US is agnostic to the $\Q[f]$ unlike EI which is used for comparison when the $\Q[f]$ is $\min[f]$.

The comparison studies show mixed results. 
For the one-dimensional function in \fref{toy_1_state}~(a) and~(b), the EKLD and US appear to converge to the true value at almost the same number of samples for inferring the statistical expectation\fref{toy_comp_1}~(a) and variance\fref{toy_comp_1}~(b) of $f$. 
However, the EKLD converges sooner for inferring $\mathbb{Q}_{2.5}[f]$.
The EI and the EKLD show similar trends on converging to the truth for numerical example no.1 when the $\Q[f]$ is the  the minimum \fref{toy_comp_1}~(c) of $f$.

For the one-dimensional function in \sref{toy_2_gen} the US and EKLD converge at around the same stage of sampling which is clearly seen in \fref{toy_comp_2}.

The three-dimensional function in \sref{toy_3_gen} is a problem with multiple dimensions. 
The EKLD converges sooner compared to the US for inferring $\E[f]$, $\V[f]$ and $\mathbb{Q}_{2.5}[f]$ when a total of 40 additional samples are collected. 
Estimating the minimum of $f$, throws up results that put the EKLD and EI at the same level of performance.

The five-dimensional problem in \sref{toy_4_gen} shows similar results as for the three-dimensional problem in the comparison study of the EKLD with US while inferring the $\E[f]$, $\V[f]$ or $\mathbb{Q}_{2.5}[f]$ of $f$.
EKLD converges sooner, near the 35 sample mark compared to the US which takes almost 50 samples to converge, for the three QoIs.
The case in \fref{toy_4_mean}~(c) and \fref{toy_comp_4}~(c)  provide similar convergence results for the EKLD and EI, with both methods converging at almost the same number of samples.

Comparison studies for the wire-drawing problem show a slightly mixed pattern of convergence with the EKLD and US taking almost same number of samples for inferring the three QoIs on which they are compared. 
Whereas for the optimization cases both EKLD and EI seem to be slow in identifying $\min[f]$ of $f$ as can be seen in \fref{wire_mean}~(d) and \fref{toy_comp_5}~(d) respectively.
Thus, results for the wire problem are not sufficient to draw a conclusion about the performance of the EKLD when compared to EI for inferring $\min[f]$. 
The next step will be to run the methodologies for more number of iterations in order to establish convergence.
\begin{figure}[!htbp]
\centering
    \subfigure[]{
        \includegraphics[width=.45\columnwidth]{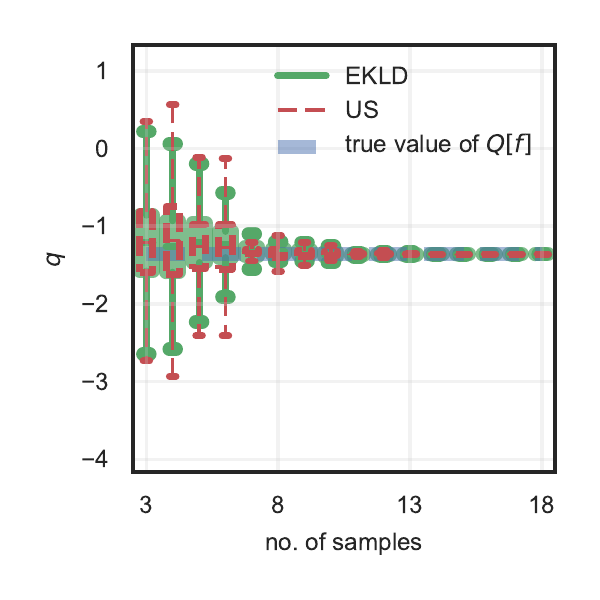}
    }
    \subfigure[]{
        \includegraphics[width=.45\columnwidth]{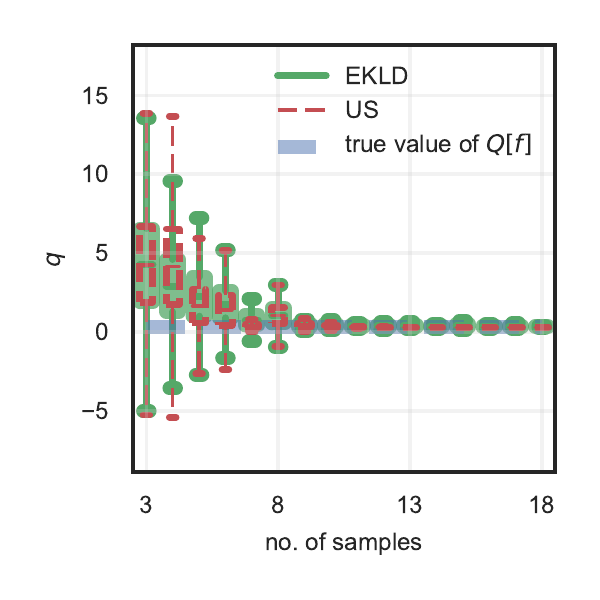}
    }
    \subfigure[]{
        \includegraphics[width=.45\columnwidth]{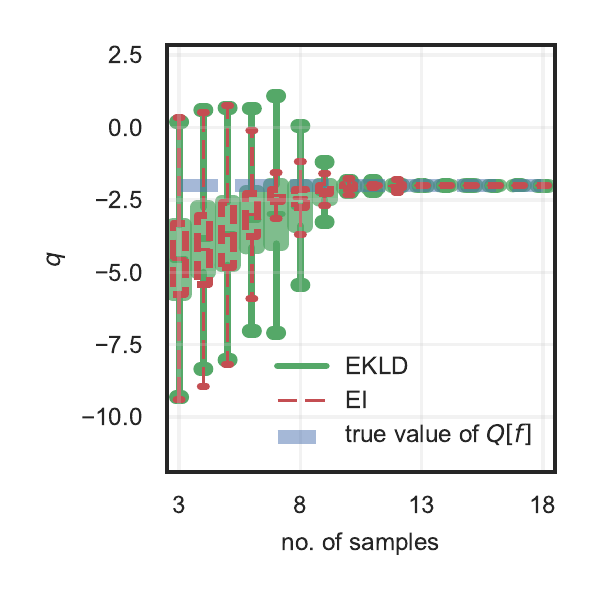}
    }
    \subfigure[]{
        \includegraphics[width=.45\columnwidth]{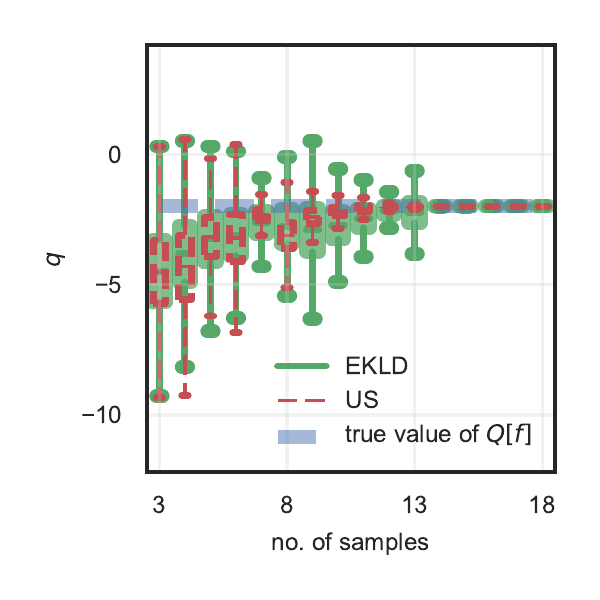}
    }
    \caption{Comparison studies for example no.1.
        Subfigures~(a), and (b), convergence of US for $\E[f]$ and $\V[f]$.
        Subfigures~(c), convergence of EI for
        $\min [f]$.
        Subfigure~(d), convergence of US for inferring $\mathbb{Q}_{2.5}[f]$.
        }
    \label{fig:toy_comp_1}
\end{figure}

\begin{figure}[!htbp]
\centering
    \subfigure[]{
        \includegraphics[width=.45\columnwidth]{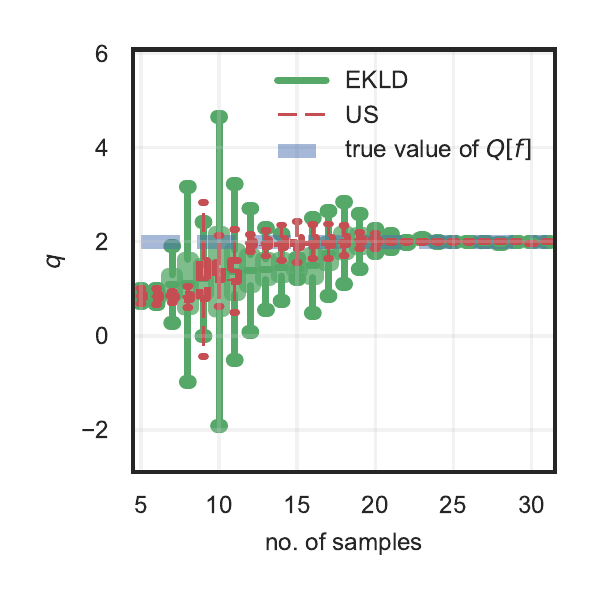}
    }
    \subfigure[]{
        \includegraphics[width=.45\columnwidth]{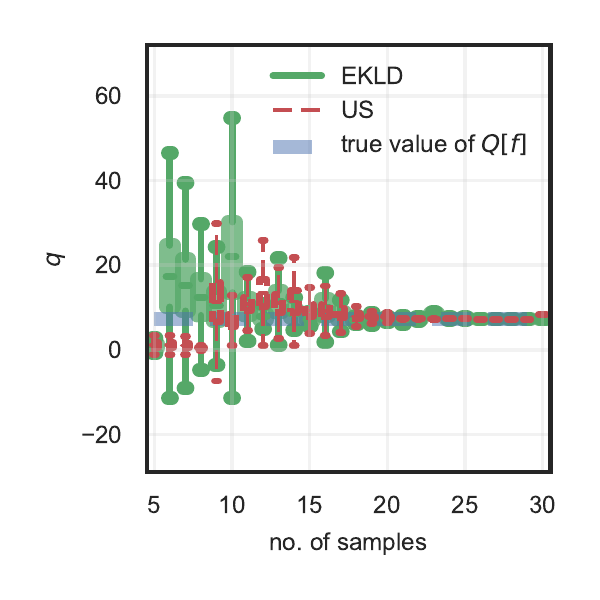}
    }
    \subfigure[]{
        \includegraphics[width=.45\columnwidth]{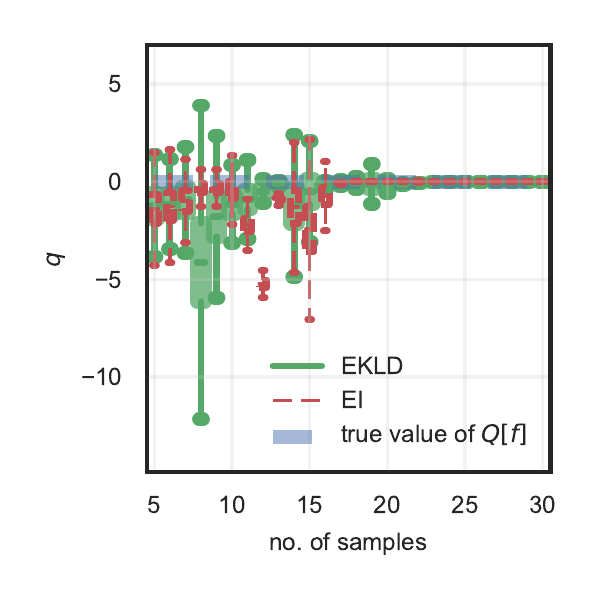}
    }
    \subfigure[]{
        \includegraphics[width=.45\columnwidth]{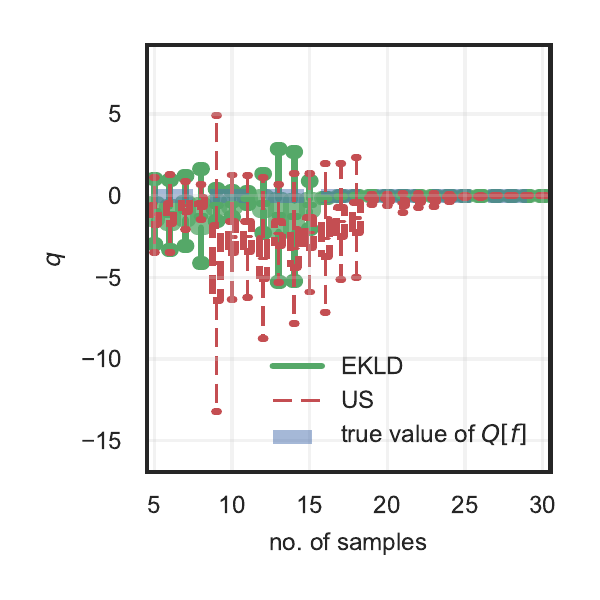}
    }
    \caption{Comparison studies for example no.2.
        Subfigures~(a), and (b), convergence of US for $\E[f]$ and $\V[f]$.
        Subfigure~(c),  convergence of EI for $\min [f]$.
        Subfigure~(d), convergence of US for inferring $\mathbb{Q}_{2.5}[f]$.
        }
    \label{fig:toy_comp_2}
\end{figure}

\begin{figure}[!htbp]
\centering
    \subfigure[]{
        \includegraphics[width=.45\columnwidth]{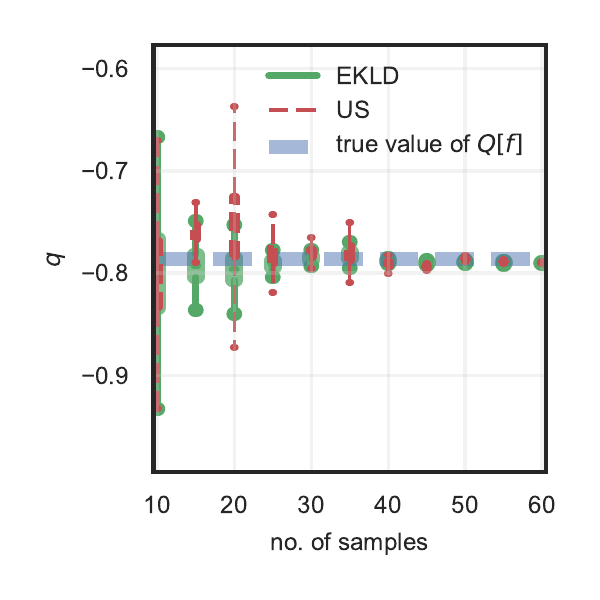}
    }
    \subfigure[]{
        \includegraphics[width=.45\columnwidth]{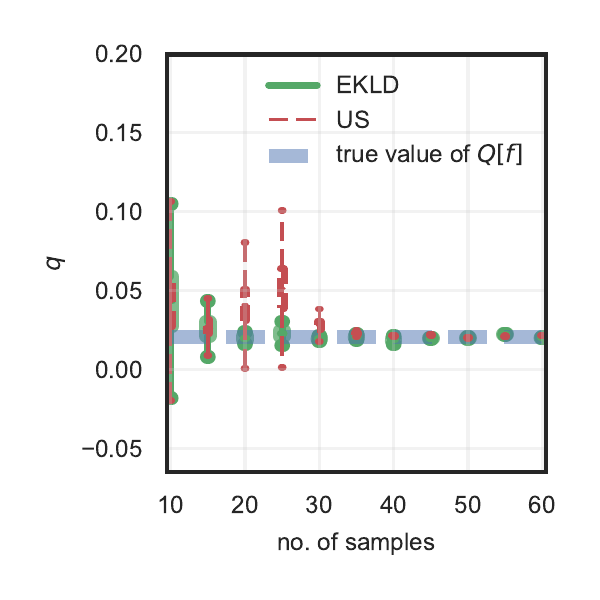}
    }
    \subfigure[]{
        \includegraphics[width=.45\columnwidth]{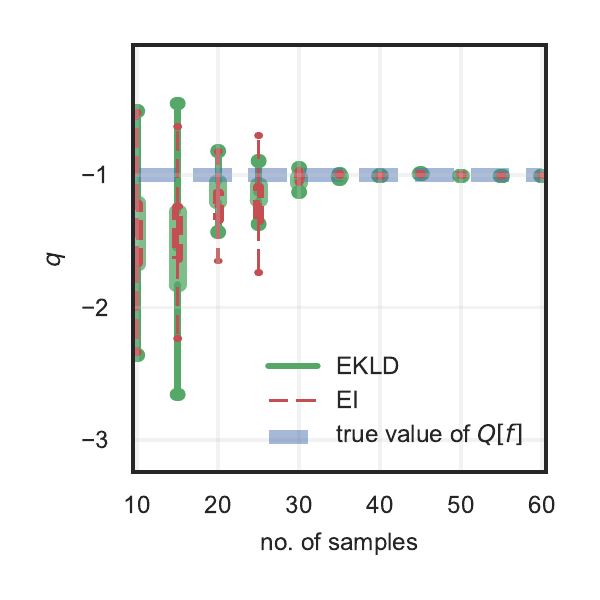}
    }
    \subfigure[]{
        \includegraphics[width=.45\columnwidth]{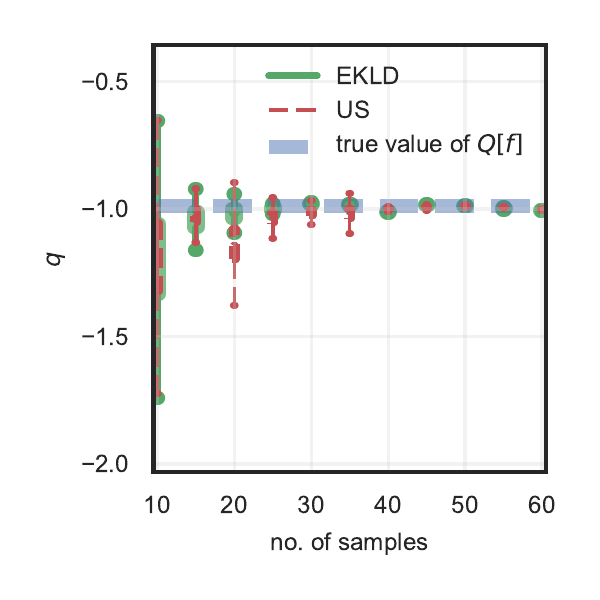}
    }
    \caption{Comparison studies for example no.3.
        Subfigures~(a), and (b), convergence of US for $\E[f]$ and $\V[f]$.
        Subfigure~(c),convergence of EI for inferring $\min [f]$.
        Subfigure~(d), convergence of US for inferring $\mathbb{Q}_{2.5}[f]$.
        }
    \label{fig:toy_comp_3}
\end{figure}

\begin{figure}[!htbp]
\centering
    \subfigure[]{
        \includegraphics[width=.45\columnwidth]{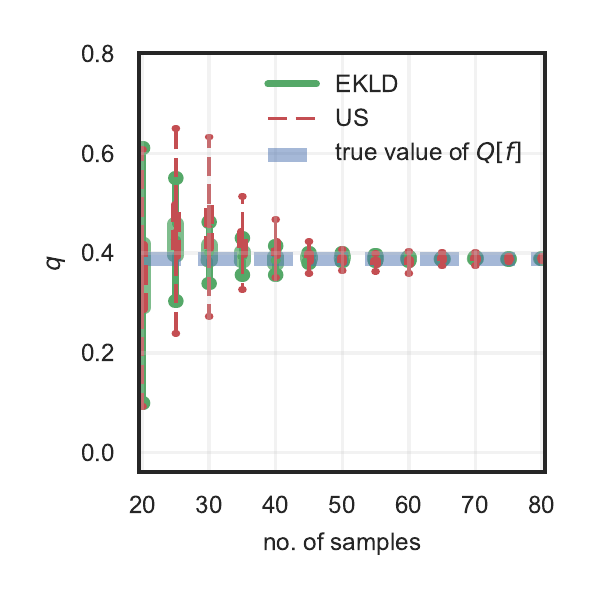}
    }
    \subfigure[]{
        \includegraphics[width=.45\columnwidth]{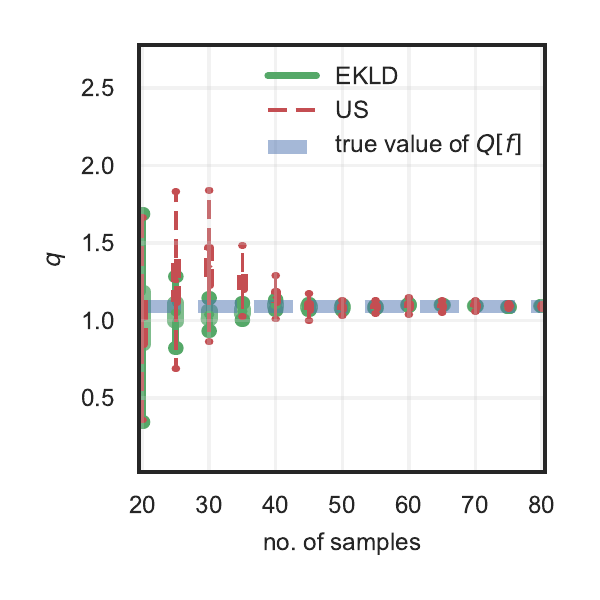}
    }
    \subfigure[]{
        \includegraphics[width=.45\columnwidth]{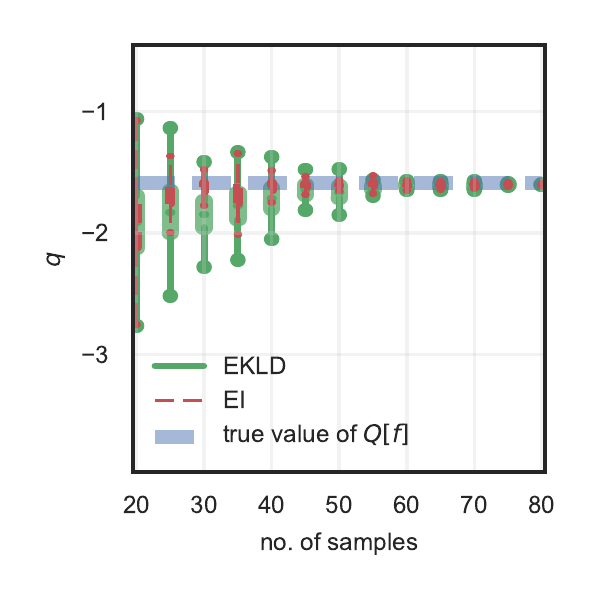}
    }
    \subfigure[]{
        \includegraphics[width=.45\columnwidth]{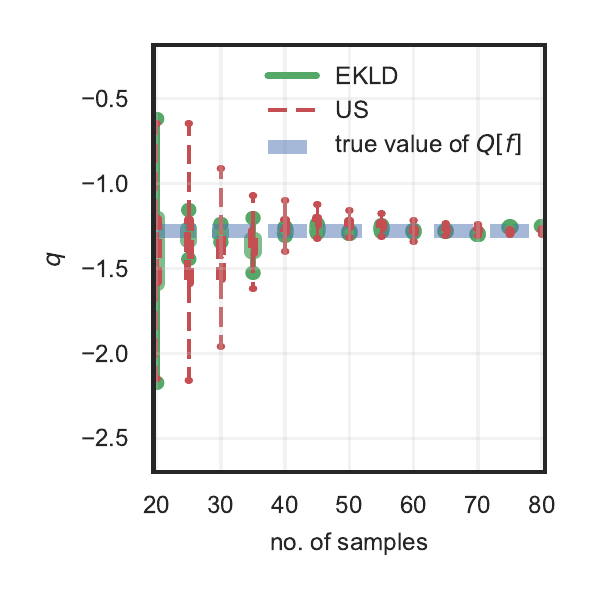}
    }
    \caption{Comparison studies for example no.4.
        Subfigures~(a), and (b), convergence of US for $\E[f]$ and $\V[f]$.
        Subfigure~(c), convergence of EI for  $\min [f]$.
        Subfigure~(e), convergence of US for inferring $\mathbb{Q}_{2.5}[f]$.
        }
    \label{fig:toy_comp_4}
\end{figure}

\begin{figure}[!htbp]
\centering
    \subfigure[]{
        \includegraphics[width=.45\columnwidth]{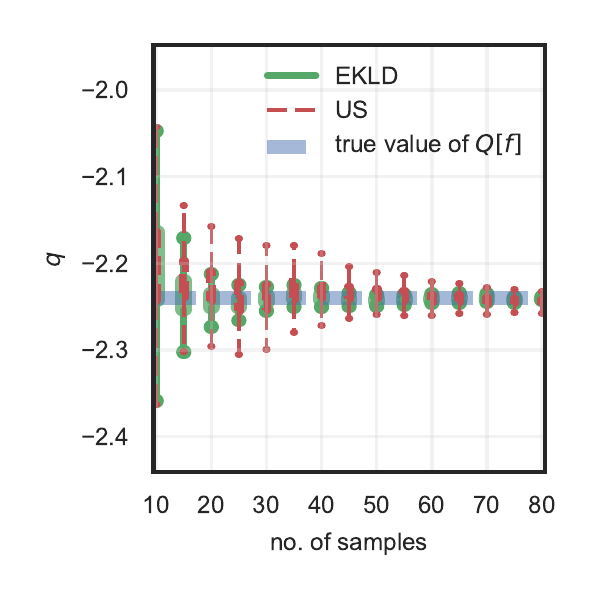}
    }
    \subfigure[]{
        \includegraphics[width=.45\columnwidth]{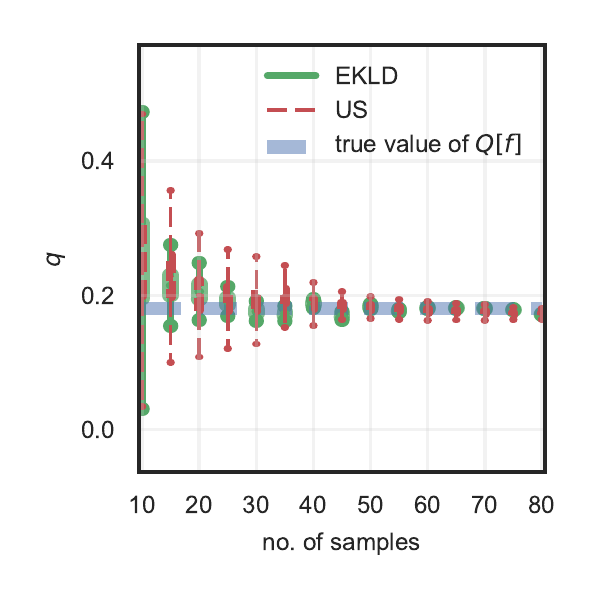}
    }
    \subfigure[]{
        \includegraphics[width=.45\columnwidth]{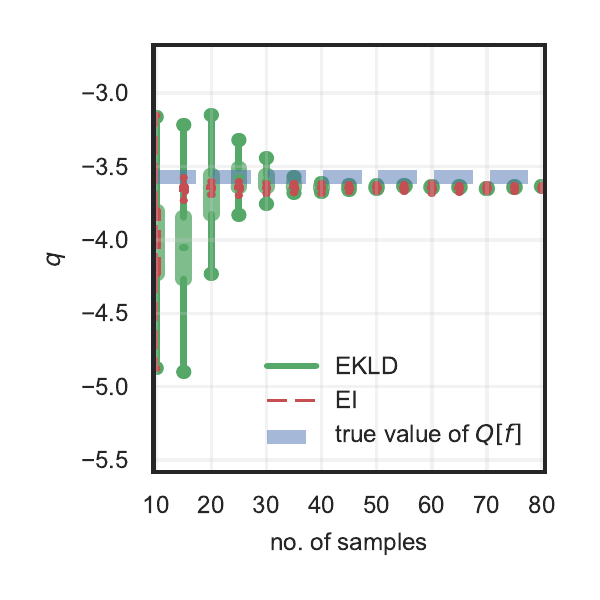}
    }
    \subfigure[]{
        \includegraphics[width=.45\columnwidth]{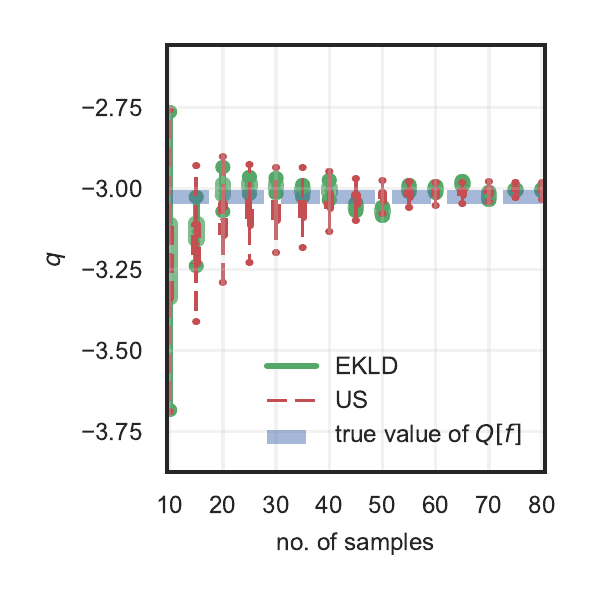}
    }
    \caption{Comparison studies for the wire-drawing problem.
        Subfigures~(a), and (b), convergence of US for $\E[f]$ and $\V[f]$.
        Subfigure~(c), convergence of EI for $\min [f]$.
        Subfigure~(d), convergence of US for inferring $\mathbb{Q}_{2.5}[f]$.
        }
    \label{fig:toy_comp_5}
\end{figure}

\section{Useful findings and insights}
\label{sec:findings}
We highlight some salient features of EKLD and its comparison studies with US and EI below.
\begin{enumerate}
    \item We derive an estimator, called EKLD, for computing information gain in a hypothetical experiment in SDOE. 
    Comparison studies demonstrate that EKLD can dynamically adapt for SDOE to infer any arbitrary $\Q[f]$.
    This is unlike the case presented in the comparison studies where the two state-of-the-art methods would perform well only in the context innate to their mathematical formulation.
    This versatile nature of EKLD stems from its analytical formulation which is an implicit function of the $\Q[f]$ being elicited.
    \item The derived estimator for EKLD samples the next point in a region of high uncertainty and/or high posterior mean of $\Q[f]$.
    This means that the EKLD, similar to the EI, balances the exploration-exploitation trade-off.
    \item The EKLD performs better or equally good when compared to US and EI for the respective $\Q[f]$s in all the case studies presented in \sref{results-bode-generic}.
    \item Some initial calibration needs to be done to select hyperpriors especially in the one-dimensional problems, where some functions can be explained better by a certain combination of hyperpriors. 
    In the numerical examples presented here we use uninformative priors for all hyperparameters. 
    A default setting for the hyperpriors has been chosen which remains the same for all problems with one input. Similarly, a default setting for the hyperpriors for problems in multiple inputs is demonstrated with promising results.
    \item The FBNSGP framework enables incorporation of point estimates of local smoothness and signal-variance even in low-sample regime. 
    For the one-dimensional numerical examples the inferred input-dependent lengthscales and signal-variances have been shown.
    The inferred values of the lengthscale and signal-variance across the input space have been sampled for each of the $S$ posterior samples of $\btheta_{m}$. This allows us to quantify the epistemic uncertainty around the point estimates of the lengthscale and the signal-variance across the input space.
    \item High input-dimensionality will pose certain challenges for the EKLD. 
    Since, training the FBNSGP model involves inferring $3$ parameters each for the lengthscale and signal-variance GPs per input dimension. 
    This means that at every stage of model training, $6d$ number of hyperparameters need to be inferred.
    In addition to this, the parameters inferred include the samples of the latent GPs at the training points which is equal to $2d$ times $N$.
    This task becomes computationally cumbersome when one is faced with problems greater than \emph{single-digit} input dimensions.
    \item An interesting point that we have not covered is the application of the EKLD framework mentioned above to suggesting multiple simulations or experiments at each iteration.
    This scheme, if extended from the current EKLD, holds great promise because this would enable practical use of computational or laboratory resources. 
    Secondly, it might also be cheaper to suggest multiple experiments in one iteration for problems in dimensions greater than five.
\end{enumerate}

\section{Conclusions}
\label{sec:conclusions_bode_generic}
We derive an estimator to quantify the information gain in a hypothetical experiment when a scientist wishes to estimate a QoI which depends on some output of the experiment.
The information gain is the Kullback-Leibler divergence between a prior state of knowledge about the QoI and a posterior state of knowledge about the QoI.
This methodology is augmented by a robust and flexible response surface modeling approach. 
The fully Bayesian non-stationary Gaussian process surrogate model allows the user to incorporate prior knowledge about the input-dependent smoothness and variance of the underlying physical response.
The performance of the SDOE heuristic is demonstrated on four numerical examples and an engineering problem of eight input dimensions.
The convergence tests for different numerical examples and the engineering problem have been compared to state-of-the-art methods namely uncertainty sampling, expected improvement and probability of improvement. 
These state-of-the-art SDOE methods are commonly suited for certain QoIs which is further highlighted by the comparison tests.
The derived SDOE heuristic converges at the same level or better as the other methods for problems which differ on accounts of dimensionality and context.
More work can be done on the presented methodology to suggest multiple experiments or designs at a single iteration, thereby allowing parallel use of laboratory or computational resources.  
This direction of research rhymes well with the spirit of batch optimization~\cite{vellanki2018bayesian} and parallel data acquisition.

\section*{Acknowledgments}
This work has been made possible by the financial support provided by National Science Foundation through
Grant 1662230.
The authors thank their collaborators at TRDDC, Tata Consultancy Services, Pune, India, 
for providing the steel wire manufacturing problem.


\bibliography{references}
\bibliographystyle{abbrv}
\end{document}